\documentclass[10pt,twocolumn,letterpaper]{article}
\usepackage{iccv}
\usepackage{times}
\usepackage{epsfig}
\usepackage{graphicx}
\usepackage{amsmath}
\usepackage{amssymb}
\usepackage{bbm}
\usepackage{booktabs}
\usepackage{multirow}
\usepackage{subfig}
\usepackage{enumitem}
\usepackage{array}
\usepackage{makecell}

\usepackage{amsmath,amsfonts,bm}

\def\vc{{\bm{c}}}

\def\vx{{\bm{x}}}

\DeclareMathAlphabet{\mathsfit}{\encodingdefault}{\sfdefault}{m}{sl}
\SetMathAlphabet{\mathsfit}{bold}{\encodingdefault}{\sfdefault}{bx}{n}

\newcommand\raiseT[2]{%
  \setbox0\hbox{$#1{#2}$}\raise\dp0\box0}

\newcommand{\bea}{\begin{eqnarray*}}
\newcommand{\eea}{\end{eqnarray*}}
\newcommand{\ba}{\begin{eqnarray*}}
\newcommand{\ea}{\end{eqnarray*}}
\newcommand{\be}{\begin{equation}}
\newcommand{\ee}{\end{equation}}
\newcommand{\bi}{\begin{itemize}}
\newcommand{\ei}{\end{itemize}}

\usepackage{array}
\newcolumntype{L}[1]{>{\raggedright\let\newline\\\arraybackslash\hspace{0pt}}m{#1}}
\newcolumntype{C}[1]{>{\centering\let\newline\\\arraybackslash\hspace{0pt}}m{#1}}
\newcolumntype{R}[1]{>{\raggedleft\let\newline\\\arraybackslash\hspace{0pt}}m{#1}}

\usepackage[pagebackref=true,breaklinks=true,letterpaper=true,colorlinks,bookmarks=false]{hyperref}

\iccvfinalcopy 

\def\iccvPaperID{6226} 

\ificcvfinal\fi

\begin{document}

\title{Boosting the Generalization Capability in Cross-Domain Few-shot Learning via Noise-enhanced Supervised Autoencoder}

\author{
Hanwen Liang$^{1}$\thanks{Equal contribution with alphabetical order. Work done when Qiong Zhang was an intern in Huawei Noah’s Ark Lab.} \thanks{Corresponding author.} \hspace{1em}
Qiong Zhang$^2$\footnotemark[1] \hspace{1em}
Peng Dai$^{1}$ \hspace{1em}
Juwei Lu$^{1}$
\\

{\tt\small hanwen.liang@huawei.com, qiong.zhang@stat.ubc.ca, \{peng.dai, juwei.lu\}@huawei.com}\\
{\small $^{1}$Huawei Noah's Ark Lab, Canada.  $^{2}$Department of Statistics, University of British Columbia, Vancouver, Canada.}
}

\maketitle
\ificcvfinal\thispagestyle{empty}\fi

\begin{abstract}
State of the art (SOTA) few-shot learning (FSL) methods suffer significant performance drop in the presence of domain differences between source and target datasets.
The strong discrimination ability on the source dataset does not necessarily translate to high classification accuracy on the target dataset.
In this work, we address this cross-domain few-shot learning (CDFSL) problem by boosting the generalization capability of the model.
Specifically, we teach the model to capture broader variations of the feature distributions with a novel noise-enhanced supervised autoencoder (NSAE).
NSAE trains the model by jointly reconstructing inputs and predicting the labels of inputs as well as their reconstructed pairs.
Theoretical analysis based on intra-class correlation (ICC) shows that the feature embeddings learned from NSAE have stronger discrimination and generalization abilities in the target domain.
We also take advantage of NSAE structure and propose a two-step fine-tuning procedure that achieves better adaption and improves classification performance in the target domain.
Extensive experiments and ablation studies are conducted to demonstrate the effectiveness of the proposed method.
Experimental results show that our proposed method consistently outperforms SOTA methods under various conditions.

\end{abstract}

\vspace{-0.2cm}
\section{Introduction}

\begin{figure}[t]
\setlength{\abovecaptionskip}{0pt}
\setlength{\belowcaptionskip}{0pt}
\centering        \includegraphics[width=0.9\linewidth]{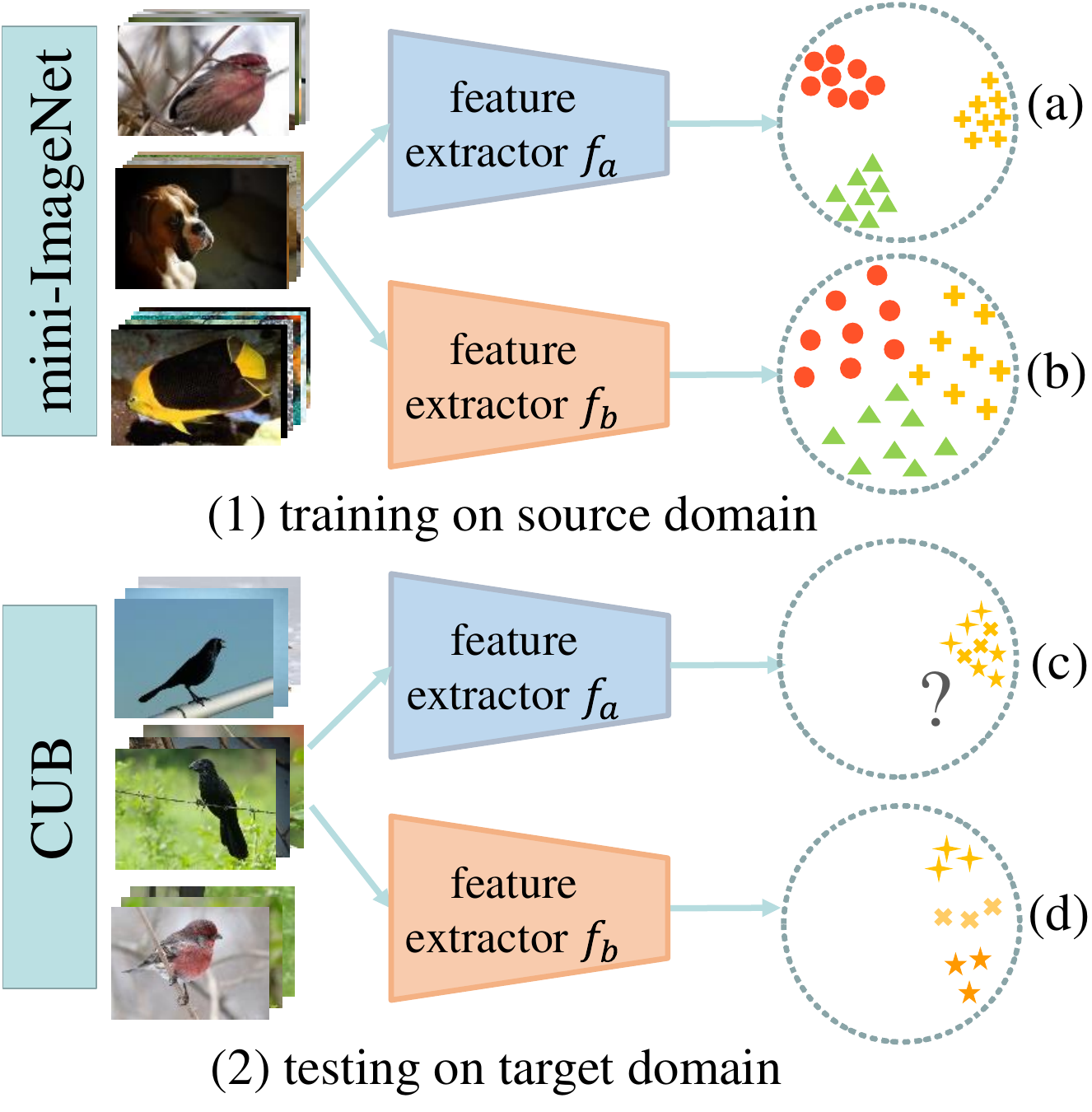}
\caption{\textbf{Motivation illustration}. Visualization of feature embeddings by a less-generalized feature extractor $f_a$  and a well-generalized feature extractor $f_b$ cross source and target domains.}
\vspace{-0.3cm}
\label{fig:intro}
\end{figure}

After years of development, deep learning methods have achieved remarkable success on visual classification tasks \cite{he2016deep, tan2019efficientnet, krizhevsky2017imagenet, quader2020weight}. The outstanding performance, however, heavily relies on large-scale labeled datasets \cite{chen2019closerfewshot}. Meanwhile, although some large-scale public datasets, e.g. ImageNet \cite{deng2009imagenet}, have made it possible to achieve better than human performance on common objects recognition, practical applications of visual classification systems usually target at categories whose samples are very difficult to collect, e.g. medical images. The scarcity of data limits the generalization of current vision systems. Therefore, it is essential to learn to generalize to novel classes with a limited number of labeled samples available in each class. 
Cross-domain few-shot learning (CDFSL) is proposed to recognize instances of novel categories in the target domain with few labeled samples. Different from general few-shot learning(FSL) where large-scale source dataset and few-shot novel dataset are from the same domain, target dataset and source dataset under CDFSL setting come from different domains, \ie the marginal distributions of features of images in two domains are quite different~\cite{zhuang2020comprehensive}.

Much work has been done to solve FSL problem and obtained promising results~\cite{vinyals2016matching,garcia2017few,snell2017prototypical,sung2018learning,finn2017model,rusu2018meta,vuorio2019multimodal}. However, ~\cite{chen2019closer, guo2020broader} show that the state-of-the-art (SOTA) meta-learning based FSL methods fail to generalize well and perform poorly under CDFSL setting. 
It is therefore of great importance to improve the generalization capability of the model and address the domain shift issue from source to target domains. ~\cite{tseng2020cross} proposes to add a feature-transformation layer to simulate various distributions of image features in training. However, this method requires access to a great amount of data from multiple domains during training.
~\cite{zhao2021domain} combines the FSL learning objective and the domain adaptation objective, while their basic assumption that source and target domain have identical label sets limits its application. ~\cite{guo2020broader} experimentally shows that the traditional transfer learning methods can outperform meta-learning FSL methods by a large margin on the benchmark. In these methods, a feature extractor is pre-trained on the source dataset and then fine-tuned on the target dataset with only a few labeled samples.
Following this thread, ~\cite{liu2020feature} proposes to regularize the eigenvalues of the image features to avoid negative knowledge transfer. 

In this work, our observation is that generalization capability plays a vital role for representation learning in cross-domain settings.
As the feature distributions of different domains are distinct, a competent feature extractor on the source domain does not necessarily lead to good performance on the target domain. It may overfit to the source domain and fail to generalize in the target domain.
Fig.~\ref{fig:intro}(a) shows an example of a less-generalized feature extractor $f_a$ that fits the source dataset very well and achieves high performance in downstream classification task. When the model is transferred to a different target domain, as shown in Fig.~\ref{fig:intro}(c), the corresponding feature embeddings of different classes may become less discriminative or even inseparable. 
On the other hand, a less perfect feature extractor $f_b$ on the source domain (Fig.~\ref{fig:intro}(b)), may have stronger generalization capability and obtain more discriminative feature embeddings in the target domain (Fig.~\ref{fig:intro}(d)). 
Under this intuition, we focus on boosting the generalization capability of the transfer learning based methods, and investigate a multi-task learning scheme that shows the potential to improve generalization performance in \cite{le2018supervised}. 
Specifically, we propose a novel noise-enhanced supervised autoencoder (NSAE) that takes more than classification tasks and learns the feature space in discriminative and generative manners.
We take advantage of the NSAE structure in the following aspects. 
First of all, it is shown in ~\cite{le2018supervised} that a supervised autoencoder can significantly improve model generalization capability. We develop the model to jointly predict the labels of inputs and reconstruct the inputs.
Secondly, motivated by the observation that ``the addition of noise to the input data of a neural network during training can, in some circumstances, lead to significant improvements in generalization performance"~\cite{reed1999neural,bishop1995training,an1996effects}, we consider reconstructed images as noisy inputs and feed them back to the system.
The joint classifications based on reconstructed and original images further improve the generalization capability and avoid the necessity of designing a mechanism to add hand-crafted noises. Thirdly, we develop a two-step fine-tuning procedure to better adapt model to the target domain. Before tuning with the supervised classification method, we first tune model on the target domain in an unsupervised manner by learning to reconstruct images in novel classes. Furthermore, theoretical analysis based on inter-class correlation (ICC) suggests that our intuition in Fig.~\ref{fig:intro} holds statistically in CDFSL settings.
Last but not the least, we claim that our proposed method can be easily added to existing transfer learning based methods to boost their performance.

Our major contributions are summarized as follows:
\begin{itemize}
    \item To the best of our knowledge, our work is the first work that proposes to use supervised autoencoder framework to boost the model generalization capability under few-shot learning settings.
    \item We propose to take reconstructed images from autoencoder as noisy inputs and let the model further predict their labels, which proves to further enhance the model generalization capability. The two-step fine-tuning procedure that does reconstruction in novel classes better adapts model to the target domain.
    \item Extensive experiments across multiple benchmark datasets, various backbone architectures, and different loss function combinations demonstrate the efficacy and robustness of our proposed framework under cross-domain few-shot learning setting. 
\end{itemize}

\section{Related work}
\noindent
\textbf{Few-Shot learning}
\label{sec:protonet}
FSL aims at recognizing examples from novel categories with a limited number of labeled samples in each class. Meta-learning scheme for FSL receives much attention for its efficiency and simplicity. Existing meta-learning based methods can be classified into two general classes: the metric-based approaches~\cite{vinyals2016matching,garcia2017few,snell2017prototypical,sung2018learning} that classify query images based on the similarity of feature embedding between query images and a few labeled images (support images), and the optimization-based approaches~\cite{finn2017model,rusu2018meta,vuorio2019multimodal} that integrate the task-specific fine-tuning and pre-training into a single optimization framework.  
However, it is shown in~\cite{chen2019closer,guo2020broader} that these SOTA methods for FSL underperform simple fine-tuning when the novel classes are from a different domain. 
Past works\cite{chi2021test,gidaris2019boosting} explore involving self-supervised learning scheme to obtain more diverse and transferable visual representations in few-shot learning. 
They fail to consider the domain-shift issue within the CDFSL settings.

\noindent
\textbf{Domain adaption}
The technique of domain adaption~\cite{wang2018deep} is usually applied to solve the domain shift issue. It aims at learning a mapping from the source domain to the target domain so that the model trained on the source domain can be applied to the target domain. However, there are some limitations of domain adaption that hinder its use in CDFSL. First, most domain adaption framework~\cite{ganin2015unsupervised,ganin2016domain,hsu2020progressive,tsai2018learning} aims at learning the mapping under the same class. For example, learn the mapping from cartoon dogs to picture of an actual dogs. This does not fit into the FSL setting where the source and target domain have different classes. There are some existing works such as~\cite{dong2018domain,li2020intelligent} that consider the domain adaption technique under FSL settings. However, these approaches require a large set of unlabeled images in the target domain, which may be very difficult or even unrealistic in practice, e.g.X-ray and fMRI images.

\noindent
\textbf{Domain generalization}
Domain generalization methods differ from domain adaption in that they aim to generalize from a set of source domains to the target domains without accessing instances from the target domain during the training stage~\cite{tseng2020cross}.
Past work to improve model generalization capability includes extracting domain-invariant features from various seen domains~\cite{muandet2013domain,blanchard2011generalizing,li2018domain}, decomposing the classifiers into domain-specific and domain-invariant components~\cite{khosla2012undoing,li2017deeper}, and augmenting the input data with adversarial learning~\cite{shankar2018generalizing,volpi2018generalizing}. However, these methods require access to multiple source domains during training. Meta-learning methods achieve domain generalization by simulating testing scenarios in source domains during training, but they perform poorly when there is a domain-shift from source domain to target domain~\cite{chen2019closer,guo2020broader}.

\noindent
\textbf{Transfer learning}
Transfer learning is a more general term for methods in which different tasks or domains are involved. One traditional transfer learning approach is the simple fine-tuning. In the simple fine-tuning, a model is trained on the source dataset and the pre-trained model is then used as initialization to train the model on the target dataset. 
It is shown in~\cite{chen2019closer,guo2020broader} that the simple fine-tuning can outperform all SOTA FSL methods under CDFSL setting.
However, when the model overfits to the source domain, the fine-tuning performs worse than directly train the same model from random initialization.
This is called negative transfer~\cite{chen2019catastrophic}. 
To avoid negative transfer and further improve the performance of simple fine-tuning under CDFSL,~\cite{liu2020feature} proposes a batch spectral regularization (BSR) mechanism by penalizing the eigenvalues of the feature matrix. 

\begin{figure*}[!htbp]
\centering
{\includegraphics[width=0.88\linewidth]{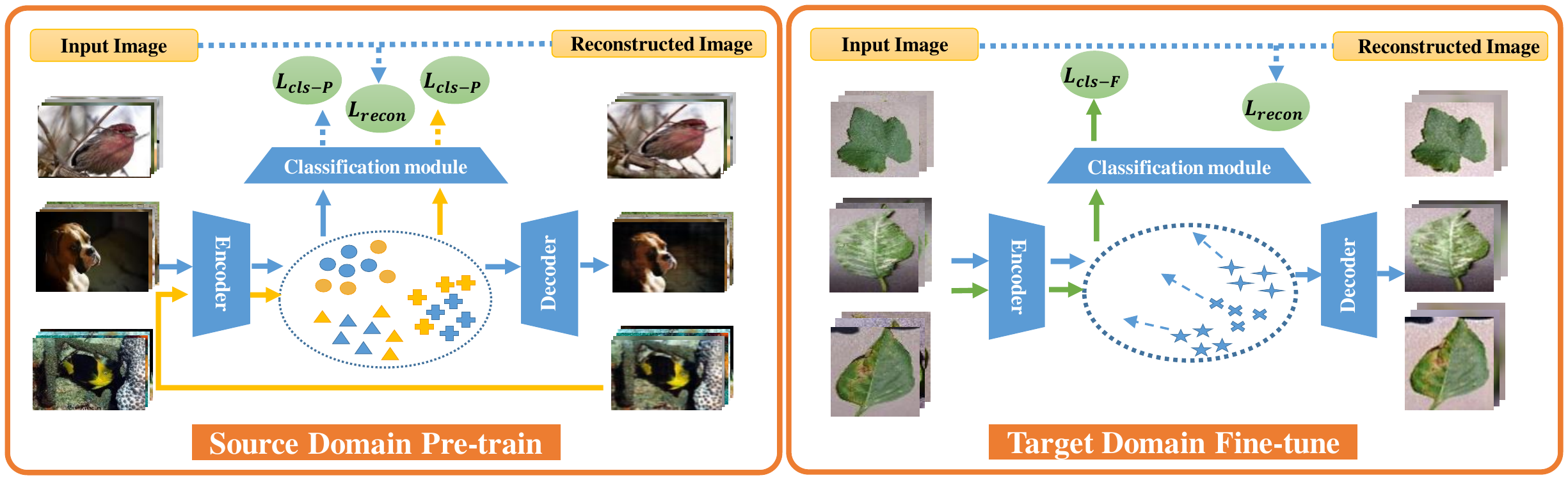}}
\caption{\textbf{An overview of the proposed pipeline.} A noise-enhanced supervised autoencoder (NSAE) is pre-trained with source dataset on the source domain to improve the generalization capability. The fine-tuning on the target domain is a two-step procedure that first performs reconstruction task on novel dataset, and then the encoder is fine-tuned for classification.}
\label{fig:pipeline}
\end{figure*}
\section{Methodology}
\subsection{Preliminaries}
\noindent
\textbf{Problem formulation}
In the cross-domain few-shot learning (CDFSL), we have a source domain $\mathcal{T}_{s}$ and a target domain $\mathcal{T}_{t}$ that have disjoint label sets. There exists a domain-shift between $\mathcal{T}_{s}$ and $\mathcal{T}_{t}$~\cite{zhuang2020comprehensive}. The source domain has a large-scale labeled dataset $\mathcal{D}_{s}$ while the target domain only has limited labeled images. Our method first pre-trains the model on the source dataset and then fine-tunes on the target dataset. Each ``N-way K-shot" classification task in target domain contains a support dataset $\mathcal{D}_{t}^{s}$ and a query dataset $\mathcal{D}_{t}^{q}$. The support set contains $N$ classes with $K$ labeled images in each class and the query set contains images from the same $N$ classes with $Q$ unlabeled images in each class. 
The goal of CDFSL is to achieve a high classification accuracy on the query set $\mathcal{D}_{t}^{q}$ when $K$ is small. 

\noindent
\textbf{Supervised autoencoder}
The autoencoder is a model that is usually used to obtain low-dimensional representations in an unsupervised manner. An autoencoder is composed of an encoder $f_{\phi}$ that encodes the input $\vx$ to its lower-dimensional representation $\tilde{\vx} = f_{\phi}(\vx)$. Then, a decoder $g_{\psi}$ decodes the representation $\tilde{\vx}$ to $\hat{\vx} = g_{\psi}(\tilde{\vx})$ which is a reconstruction of input $\vx$. The goal of the autoencoder is to minimize the difference between the input $\vx$ and its reconstruction $\hat{\vx}$ and the reconstruction loss is formulated as
\begin{equation}\label{eq:obj_autoencoder}
    \mathcal{L}_{\text{REC}}(\phi,\psi;\vx) = \|\vx-\hat{\vx}\|_2. 
\end{equation}

When the labels of the inputs are available, the supervised autoencoder (SAE)~\cite{le2018supervised} that jointly predicts the class label and reconstructs the input is proved to generalize well for downstream tasks. In the SAE, the representation $\tilde{\vx}$ is fed into a classification module for label prediction and the loss function is
\begin{equation}
\label{eq:obj_SAE}
    \mathcal{L}_{\text{SAE}}^{\lambda,cls}(\phi,\psi;\vx,y) =
    \mathcal{L}_{cls}(\tilde{\vx},y) + \lambda \mathcal{L}_{\text{REC}}(\phi,\psi;\vx)
\end{equation}
where $\mathcal{L}_{cls}$ is a loss function for classification and $\lambda$ is a hyper-parameter that controls the reconstruction weight.

\subsection{Overview}
Under CDFSL setting,~\cite{guo2020broader} shows that the traditional transfer learning based methods outperform all FSL methods. In the traditional transfer learning based method, a feature extractor is first pre-trained on the $\mathcal{D}_{s}$ with sufficient labeled images by minimizing the classification loss $\mathcal{L}_{\text{cls-P}}$. Then the pre-trained feature extractor is fine-tuned on the target domain support set $\mathcal{D}_{t}^{s}$ by minimizing the classification loss $\mathcal{L}_{\text{cls-F}}$. Note that the loss functions $\mathcal{L}_{\text{cls-P}}$ during the pre-training and $\mathcal{L}_{\text{cls-F}}$ during the fine-tuning may be different. 
Considering the superior performance of traditional transfer learning method on CDFSL, we use the transfer learning pipeline in our work. Motivated by the generalization capability of SAE and the generalization enhancement by feeding noisy inputs, we propose to boost the generalization capability of model via a noise-enhanced SAE (NSAE). To achieve this, we train a SAE that learns the feature space in generative and discriminative manners. NSAE not only predicts the class labels of the inputs but also predicts the labels of the ``noisy" reconstructions. 
We also leverage the NSAE to perform domain adaption during the fine-tuning. Specifically, it is tuned to reconstruct target domain images before tuned to do classification task. An overview of our proposed pipeline is depicted in Fig.~\ref{fig:pipeline}. A detailed explanation is given in the following sections.

\subsection{Pre-train on the source domain}
To borrow information from the source domain, the first step is to train a feature encoder on the large-scale source domain.
Instead of training a single feature encoder, we propose to train a NSAE on the source domain. 
Let $\mathcal{D}_{s} = \{(\vx_{m}^{s}, y_{m}^{s}), m=1,2,\cdots,M\}$ be the source dataset where $M$ denotes the number of classes, and let $f_{\phi}$ and $g_{\psi}$ be the encoder and decoder respectively. The input images are fed into $f_{\phi}$ to extract the feature representations which are fed into $g_{\psi}$ to reconstruct the original inputs. Meanwhile, the feature representations are also fed into a classification module to predict the class labels of inputs. In our formulation, the reconstructed images are seen as ``noisy" inputs which are further fed back into the encoder for classification.
The NASE is trained to reconstruct the input images and predict the class labels of both original and reconstructed images.
The loss function of NSAE during the pre-training is
\begin{equation}
\label{eq:pre-train_loss}
\begin{split}
\mathcal{L}_{\text{NSAE}}(\phi,\psi;\mathcal{D}_s)=&\frac{1}{M}\sum_{m=1}^{M} \mathcal{L}_{\text{SAE}}^{\lambda_1,\text{cls-P}}(\phi,\psi;\vx_{m}^s, y_{m}^s)\\
&+\frac{\lambda_2}{M}\sum_{m=1}^{M}\mathcal{L}_{\text{cls-P}}(\boldsymbol\theta;f_{\phi}(\hat{\vx}_{m}^s), y_{m}^s)
\end{split}
\end{equation}
where $\mathcal{L}_{\text{cls-P}}$ is some classification loss and $\mathcal{L}_{\text{SAE}}$ is given in~\eqref{eq:obj_SAE}. The second term is the classification loss of reconstructed images. The classification loss functions for the original inputs and the reconstructed images are the same. $\lambda_1$ and $\lambda_2$ are two hyper-parameters that control the weights of losses.

We show in the ablation study that the use of the classification loss based on the reconstructed images is indispensable which further improves the generalization capability of the feature encoder.

\subsection{Fine-tune on the target domain}
\label{sec:fine-tune}
The second stage is to fine-tune the pre-trained model on the target domain where only a very limited number of labeled examples are available.
Based on the nature of our autoencoder architecture, we propose a two-step procedure for domain adaptation to the target domain. 

Let $D_{t}^{s}=\{(\vx_{ij},y_{ij});i=1,2,\ldots, N, j=1,\ldots,K\}$ be the support set on the target domain. In the first step, we leverage the autoencoder architecture and propose to perform domain adaption by reconstructing the support images for certain epochs. The model aims at minimizing reconstruction loss $\sum_{i,j}\mathcal{L}_{\text{REC}}(\phi,\psi;\vx_{ij})$. 
In the second step of the fine-tuning, only the encoder is used to fine-tune on $\mathcal{D}_{t}^{s}$ with the classification loss $\mathcal{L}_{\text{cls-F}}$.
We show in the ablation study that such a two-step procedure works better than purely fine-tuning the encoder with $\mathcal{L}_{\text{cls-F}}$ on the target support set. We refer to them as one-step or two-step fine-tuning respectively in the following.

In traditional fine-tuning, all the parameters of the encoder or the first several layers of the encoder are fixed when the parameters of the classification module are updated. However, ~\cite{guo2020broader} shows that, under the CDFSL setting, the fine-tuned model can achieve better performance when the model is completely flexible. Therefore, we update all parameters of the model during the fine-tuning stage.

\subsection{Choices of loss functions}
\label{sec:loss-function}
The loss functions $\mathcal{L}_{\text{cls-P}}$ and $\mathcal{L}_{\text{cls-F}}$ are not specified in the description above. 
In fact, they can be any sensible loss functions for classification. 
In this paper, we study two loss functions for $\mathcal{L}_{\text{cls-P}}$ in pre-training stage, the first one is the cross entropy (CE) loss
\begin{equation}
    \mathcal{L}_{\text{CE}}(\mathbf{W};\vx,y) =-\log\left\{\frac{\exp((\mathbf{W}\vx)_{y})}{\sum_{c} \exp((\mathbf{W}\vx)_{c})}\right\}
\end{equation}
where $\mathbf{W}$ is the parameters of the linear classifier and $(\cdot)_{c}$ means the $c$th element of the corresponding vector.
The second one is the CE loss with batch spectral regularization (BSR)~\cite{liu2020feature} that regularizes the singular values of the feature matrix in a batch. This classification loss is referred to as BSR loss and is given by
\begin{equation}
\label{eq:bsr-loss}
    \mathcal{L}_{\text{BSR}}(\mathbf{W}) = \mathcal{L}_{\text{CE}}(\mathbf{W}) + \lambda \sum_{i=1}^b \sigma_i^2
\end{equation}
where $\sigma_1,\sigma_2,\ldots,\sigma_b$ are singular values of the batch feature matrix.

In the second step of the fine-tuning stage, we consider the traditional fine-tuning and the distance-based fine-tuning. 
In traditional fine-tuning method, a linear classifier on top of the feature extractor is fine-tuned to minimize the CE loss.
In the distance-based fine-tuning method, we follow the simple but effective distance-based classification method~\cite{snell2017prototypical} in FSL, where the images are classified based on their similarities to the support images. 
To use distance-based loss function during the fine-tuning, at each iteration of the optimization, within each class of $\mathcal{D}_{t}^{s}$, we randomly split half of the images into a pseudo-support set $\mathcal{D}_{t}^{ps}=\{(\vx_{ij}^{s}, y_{ij}^{s}),i=1,2,\ldots,N,j=1,2,\ldots,K/2\}$ and the rest to a pseudo-query set $\mathcal{D}_{t}^{pq} =\{(\vx_{ij}^{q}, y_{ij}^{q}),i=1,2,\ldots,N,j=1,2,\ldots,K/2\}$. The feature embeddings of the pseudo-support set and the pseudo-query set based on the the feature extractor $f_{\phi}$ is first obtained. Then the mean feature embeddings of the pseudo-support images in the same class
\begin{equation}
\vc_{i} = \frac{K}{2}\sum_{j=1}^{K/2} f_{\phi}(\vx_{ij}),~i=1,2,\ldots,N
\end{equation}
is used to represent the class and is called the class prototype. 
Given a distance function $d(\cdot,\cdot)$ and a pseudo-query image $\vx$, the classification module produces a distribution over classes. 
The probability that $\vx$ belongs to class $k$ is given as:
\begin{equation}
\label{eq:distance_prob}
    \mathbb{P}(y=k|\vx) = \frac{\exp(-d(f_{\phi}(\vx),\vc_k))}{\sum_{k'}\exp(-d(f_{\phi}(\vx),\vc_{k'}))}
\end{equation}
Since the true class labels of the pseudo-query images are known, the parameter $\phi$ can therefore be fine-tuned by maximizing the log-likelihood of the images in the query set, that is
\begin{equation}
\label{eq:protonet-loss}
    \mathcal{L}_{\text{D}}(\phi) = \sum_{i,j}\log \mathbb{P}(y=y_{ij}^q|\vx_{ij}^q).
\end{equation}
It is shown in~\cite{snell2017prototypical} that the distance-based classifier is effective. After the feature encoder is fine-tuned with the classification loss, we use the full support set to build the class prototypes and then classifies the query image into the class that has the highest probability in~\eqref{eq:distance_prob}. This is equivalent to classify the query image with the nearest neighbor classifier, the query image is classified to class $k$ if it is closest to $k$th class prototype.
We use cosine distance for $d(\cdot,\cdot)$ in our experiment.

The combination of the two loss functions for $\mathcal{L}_{\text{cls-P}}$ and the two loss functions for $\mathcal{L}_{\text{cls-F}}$ leads to $4$ different loss functions respectively named as CE+CE, BSR+CE, CE+D, and BSR+D. The first acronym is referring to the loss function for $\mathcal{L}_{\text{cls-P}}$ and the second acronym is referring to the loss function for $\mathcal{L}_{\text{cls-F}}$. 

\section{Experiments}
In this section, we demonstrate the efficacy of our proposed method for CDFSL on benchmark datasets via extensive experiments and ablation studies.  

\subsection{Experiment setting}
\label{sec:exp_set}
\noindent
\textbf{Dataset}
Following the benchmark~\cite{guo2020broader}, we use \textbf{\emph{miniImageNet}} as the source dataset, which is a subset of ILSVRC-2012~\cite{russakovsky2015imagenet}. 
It contains $100$ classes with $600$ images in each class. 
Following the convention, the first $64$ classes are used as the source domain images to pre-train the model in our experiment. 
To evaluate the generalization capability of our method, we use 8 different datasets as the target domains. 
The first four datasets are the benchmark datasets proposed in~\cite{guo2020broader}. 
We refer to these four dataset as~\textbf{\emph{CropDisease}}, \textbf{\emph{EuroSAT}}, \textbf{\emph{ISIC}}, \textbf{\emph{ChestX}} in the following, and the similarity of these datasets to mini-ImageNet decreases from left to right. 
We also include another four natural image datasets,~\textbf{\emph{Car}}~\cite{KrauseStarkDengFei-Fei_3DRR2013},~\textbf{\emph{CUB}}~\cite{branson2010visual,welinder2010online},~\textbf{\emph{Plantae}}~\cite{van2018inaturalist}, and \textbf{\emph{Places}}~\cite{zhou2017places} that are commonly used in CDFSL~\cite{tseng2020cross}.

\noindent
\textbf{Evaluation protocol}
To make a fair comparison with existing methods for CDFSL, we evaluate the performance of the classifiers by simulating $600$ independent $5$-way few-shot classification tasks on each target domain dataset. 
For each task, we randomly sample $5$ classes and within each class, we randomly select $K$ images as the support set and $15$ images as the query set. Following the benchmark~\cite{guo2020broader}, we let $K=5, 20, 50$. 
In 50-shot classification, the \emph{Car} dataset has only a few classes that have more than 50 images, so we do not consider this dataset; the \emph{CUB} dataset has 144 out of 200 classes that have more than 60 images per class, so we sample from these 144 classes and use $10$ images per class as query set for $5$-way $50$-shot evaluation.
Then for each task, we fine-tune the pre-trained model on the support set and evaluate its performance on the query set. 
Transductive inference~\cite{guo2020broader,nichol2018first} is used that the statistics of the query images are used in batch normalization.
In total, the pre-trained model is fine-tuned and evaluated for $600$ times under each experiment setting, and the average classification accuracy as well as 95\% confidence interval on the query set is reported.

\noindent
\textbf{Network architecture}
To illustrate the effectiveness of the supervised autoencoder, we consider two commonly used encoder architectures in the experiment, namely Conv4~\cite{vinyals2016matching} and ResNet10~\cite{guo2020broader}. 
Besides the difference in network architecture, these two networks have different input sizes. 
We resize the source and target domain images to $84\times84$ for Conv4 and $224\times 224$ for ResNet10.
We design different decoder architectures for these two encoders. 
The decoders we designed consist of deconvolutional blocks, with each block containing 2D transposed convolution operator and ReLU activation, which expand the dimension of the feature map.
To mirror the dimension of the output in the encoder, we set the hyperparameters in the 2D transposed convolution layer to be $kernel size = 2$ and $stride=2$. 
The architecture and layer specifications of the autoencoder can be found in Section B of the supplementary material.

\noindent
\textbf{Hyper-parameter settings} 
All of our experiments are conducted in~\emph{pytorch}~\cite{paszke2019pytorch}. 
We use the same set of optimizers and hyper-parameters for all experiments regardless of model architecture and the target domain.
Specially, in the pre-training stage, the model is trained from scratch, with a batch size of $64$ for $400$ epochs. 
We use combinations of random crop, random flip, and color jitter to augment the source dataset. We let $\lambda_1=\lambda_2=1$ in~\eqref{eq:pre-train_loss} and let $\lambda=0.001$ in ~\eqref{eq:bsr-loss}. We optimize our model with stochastic gradient descent(SGD), with a learning rate of $10^{-3}$, the momentum of $0.9$, and weight decay of $5\times 10^{-4}$. 
In the fine-tuning stage, we also use SGD optimization. In the first step, we use a learning rate of $10^{-3}$ and do reconstruction task for 30 epochs. In the second step, we use a learning rate of $10^{-2}$, the momentum of $0.9$, and weight decay of $10^{-3}$ and fine-tune for 200 epochs. In the distance-based fine-tuning, as pointed in Section \ref{sec:loss-function}, half of the support set is used as pseudo-support set and the other half is used as pseudo-query set. In the traditional fine-tuning, the batch size of 4 is used for $5$ and $20$ shot, and 16 for $50$ shot.

\noindent
\textbf{Data augmentation \& label propagation}
A simple but effective way in FSL is to supplement the small support set with hand-crafted data augmentation~\cite{liu2020feature}.
The operations such as random crop, random flip, and color jitter can be used to augment the dataset. 
We use the same combination of operations as shown in Table 1 in~\cite{liu2020feature} when the data augmentation technique is used during the fine-tuning. 
For the distance-based fine-tuning, the order of data split and data augmentation leads to a difference in the training set. In our experiment, we first augment the support set and then randomly split the augmented images within the same class into a pseudo-support set and a pseudo-query set. 
For the traditional fine-tuning, we augment the support set and at each iteration of the fine-tuning, a random batch is selected to compute the gradient. 
To further improve the classification accuracy, a post-processing method called label propagation~\cite{liu2020feature} is also used in our method. 
The label propagation refines the predicted labels based on the similarities within the unlabeled query set.

\begin{figure}[!pb]
\centering
\subfloat[Conv4]{\includegraphics[width=0.5\linewidth]{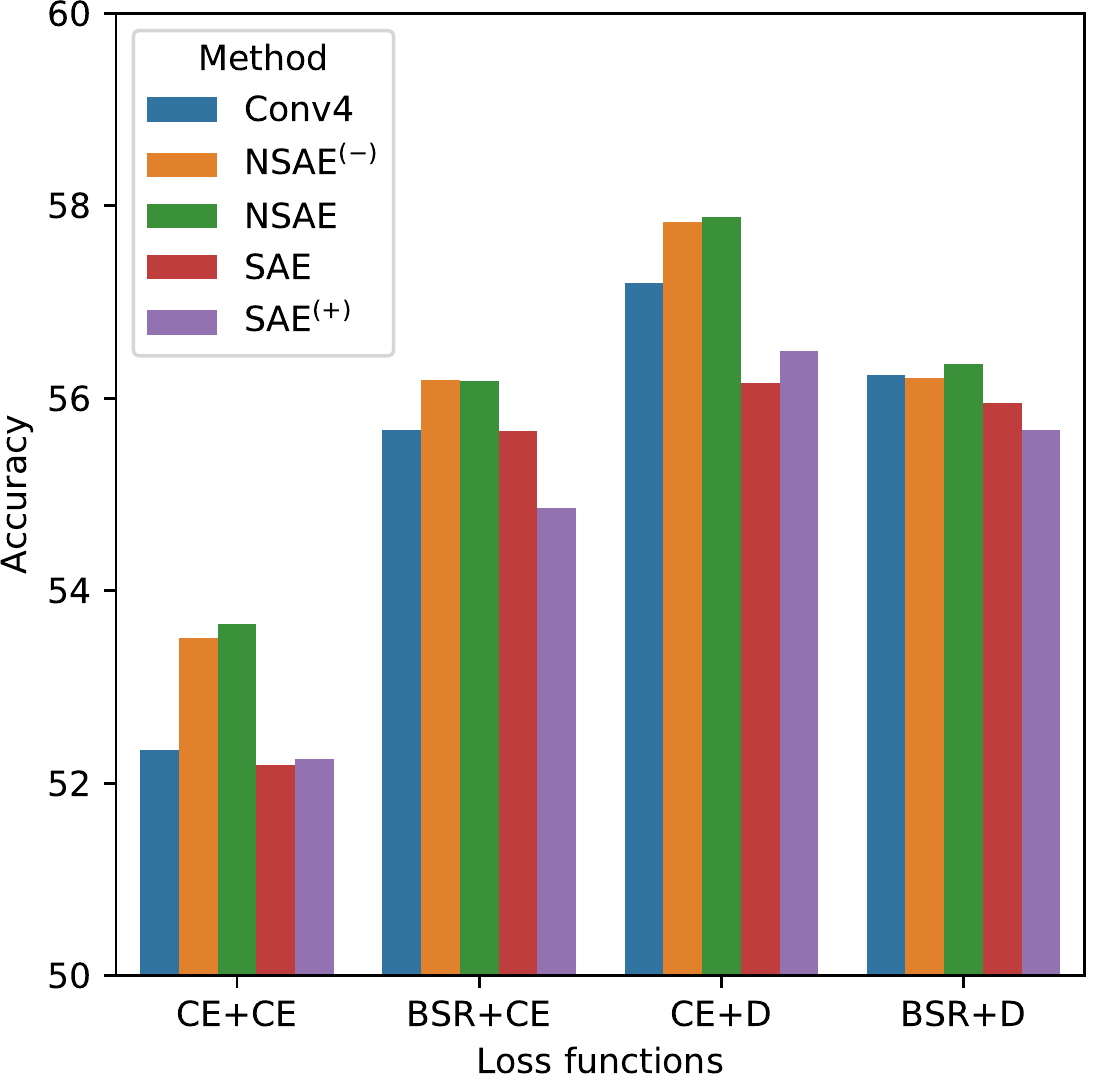}}
\subfloat[ResNet10]{\includegraphics[width=0.5\linewidth]{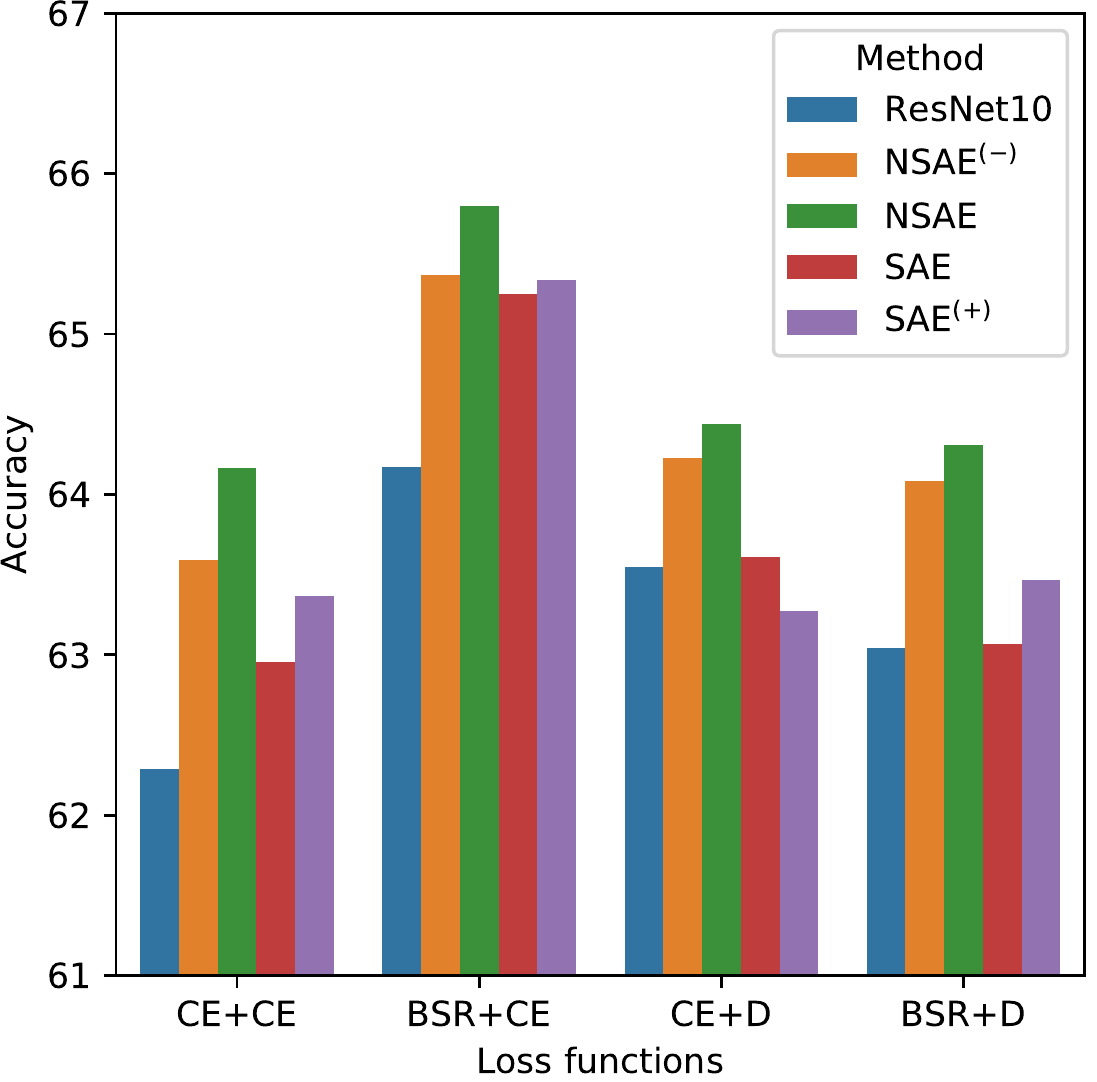}}
\caption{\textbf{Ablation study visualization.} The average 5-way 5-shot classification accuracy over 8 datasets when encoder is (a) Conv4 and (b) ResNet10. Within each plot, the bars are grouped by the classification loss functions during pre-training and fine-tuning on x-axis. Our proposed method NSAE is represented by the green bar.}
\label{fig:ablation}
\end{figure}

\subsection{Ablation study}
To study the effectiveness of our proposed method, we conduct an ablation study under $5$-way $5$-shot setting on all 8 datasets with different architectures to show that
\begin{enumerate}[label={(\arabic*)}]
    \item
    \label{itm:proposal_better}
    with the same classification loss combination, our proposed method boosts generalization capability and obtains consistently better performance on the target domain than the traditional transfer learning based methods for CDFSL;
    \item 
    \label{itm:noise}
    the noise-enhancement that predicts class labels of reconstructed images is necessary and can greatly improve the generalization capability during pre-training;
    \item 
    \label{itm:2step}
    the proposed two-step fine-tuning procedure achieves better domain adaption and leads to higher classification accuracy than the traditional one-step fine-tuning.
\end{enumerate}

The ablation study is conducted using four kinds of combinations of the classification loss functions for pre-training and fine-tuning, i.e. CE+CE, BSR+CE, CE+D, and BSR+D.
The average classification accuracy across 8 datasets is visualized in Fig.~\ref{fig:ablation}. 
The results based on two different encoder architectures, i.e. Conv4 and ResNet10, are respectively give in Fig.~\ref{fig:ablation} (a) and Fig.~\ref{fig:ablation} (b). The details of results of 8 datasets under different settings can be found in Section C in supplementary file.

To show~\ref{itm:proposal_better}, we compute the 5-way 5-shot classification accuracy when we only train a single encoder on the source domain and when we train a NSAE on the source domain. These two cases are labeled as ResNet10 (Conv4) and NSAE in Fig.~\ref{fig:ablation}. As is shown in the plot, our proposed method always has higher classification accuracy regardless of the encoder architecture and the classification functions.

To show~\ref{itm:noise}, we compare our proposed method with two extreme cases. The first case is the SAE where we do not further feed in the reconstructed images for classification during the pre-training. The second case is the SAE$^{(*)}$ where we double the weight on the classification loss of original images as if the auto-encoder works perfectly that the reconstructed images are identical to original images. 
As shown in the figure, the NSAE surpasses the other two variants under different settings. It suggests that the classification loss based on the reconstructed images is necessary. Without this loss, the two extreme cases that we compare with could even be worse than the traditional transfer learning based methods.
We also compare our proposed method with that when hand-crafted noisy images are used, the results can be found in the supplementary material Section D.

To show~\ref{itm:2step}, we train two NSAEs with the same pre-training method and fine-tune the pre-trained autoencoder either with a one-step procedure or a two-step procedure as described in Section~\ref{sec:fine-tune}.
These two cases are respectively named as NSAE(-) and NSAE. It is shown in the figure that the two-step fine-tune procedure also outperforms the one-step fine-tune procedure.

From the ablation study, we can also see that when the loss functions during the pre-training and fine-tuning are the same, the more complex encoder ResNet10 gives a higher classification accuracy compared with Conv4. When the pre-training classification loss is CE, using the distance-based loss function during the fine-tuning gives higher classification accuracy than using CE loss. However, when using classification loss BSR during pre-training, we get an opposite conclusion. 
Overall, using BSR as classification loss during pre-training and CE as classification loss during fine-tuning achieves the highest accuracy.

\begin{figure}[!htbp]
\centering
\begin{tabular}{@{}c@{}@{}c@{}@{}c@{}@{}c@{}}
\includegraphics[width=0.248\linewidth]{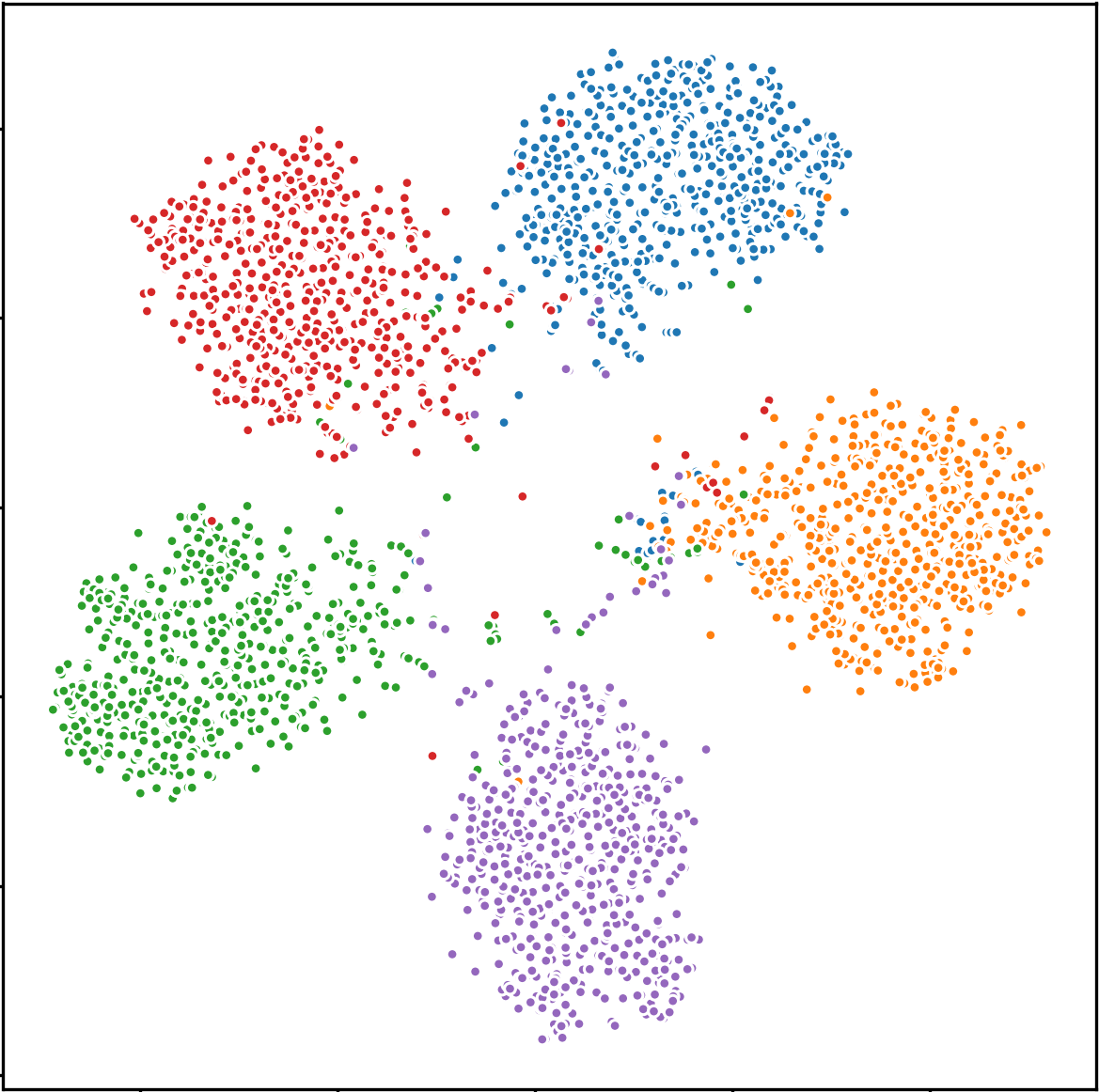}&
\includegraphics[width=0.248\linewidth]{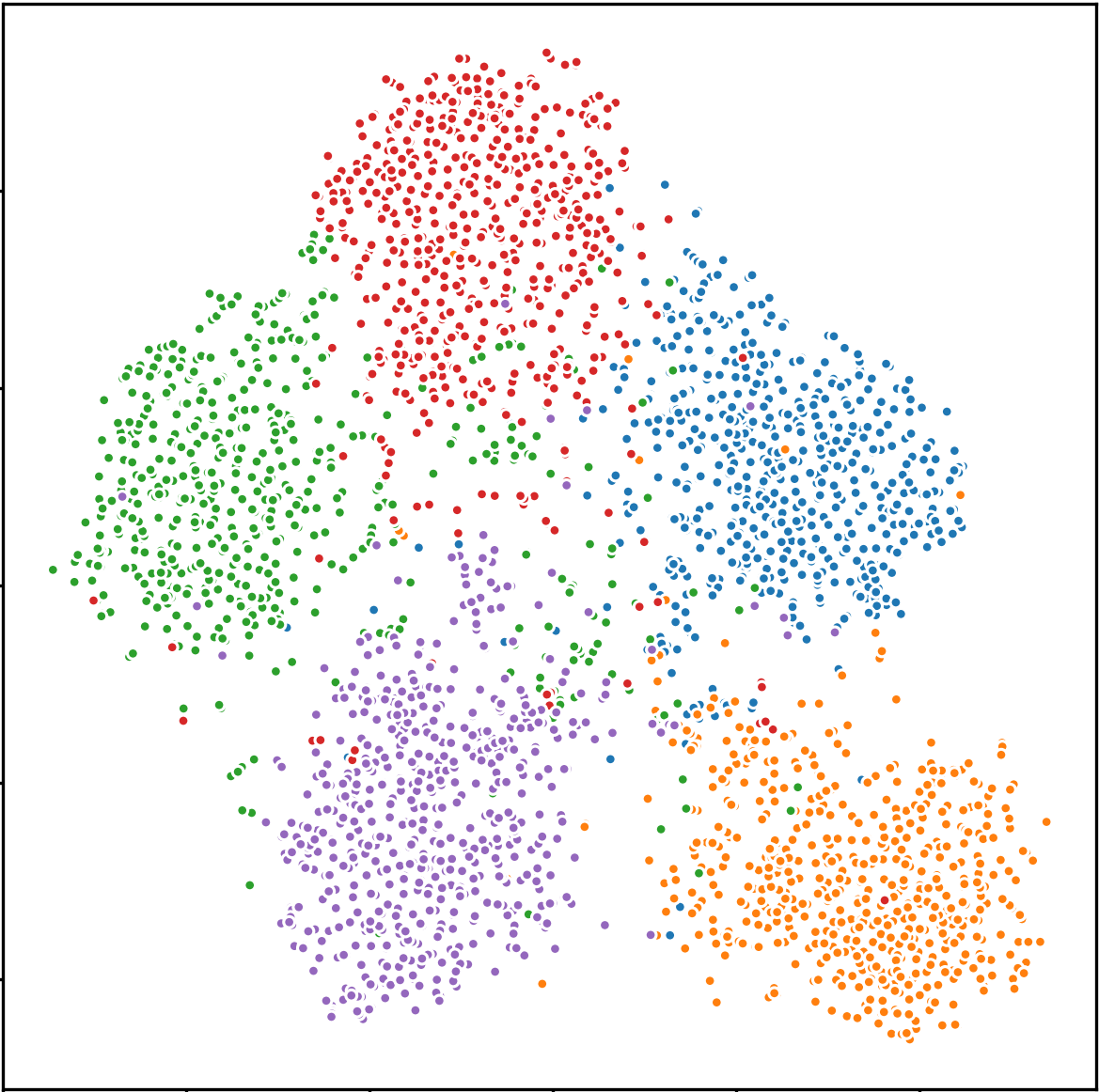}&
\includegraphics[width=0.248\linewidth]{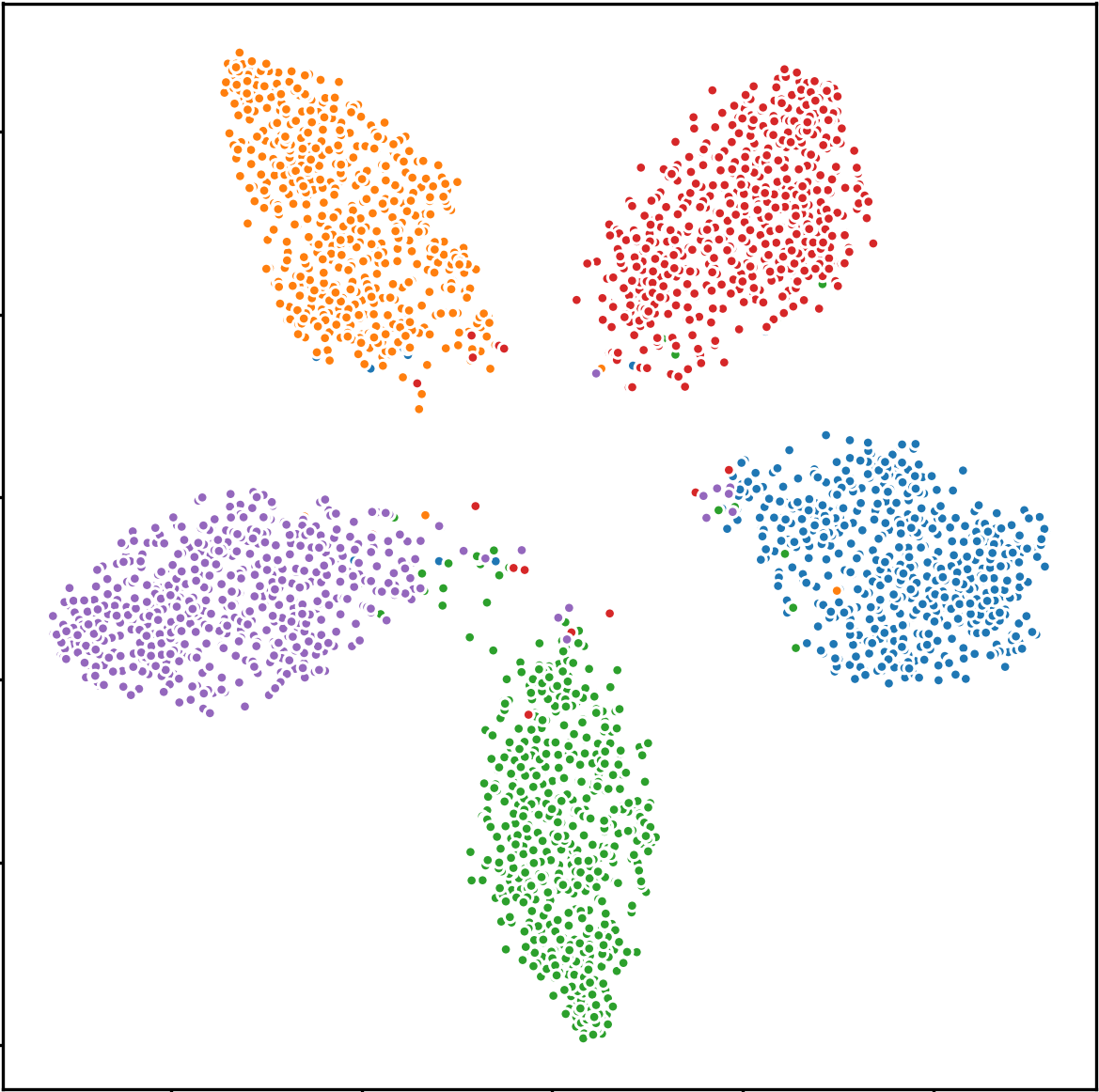}&
\includegraphics[width=0.248\linewidth]{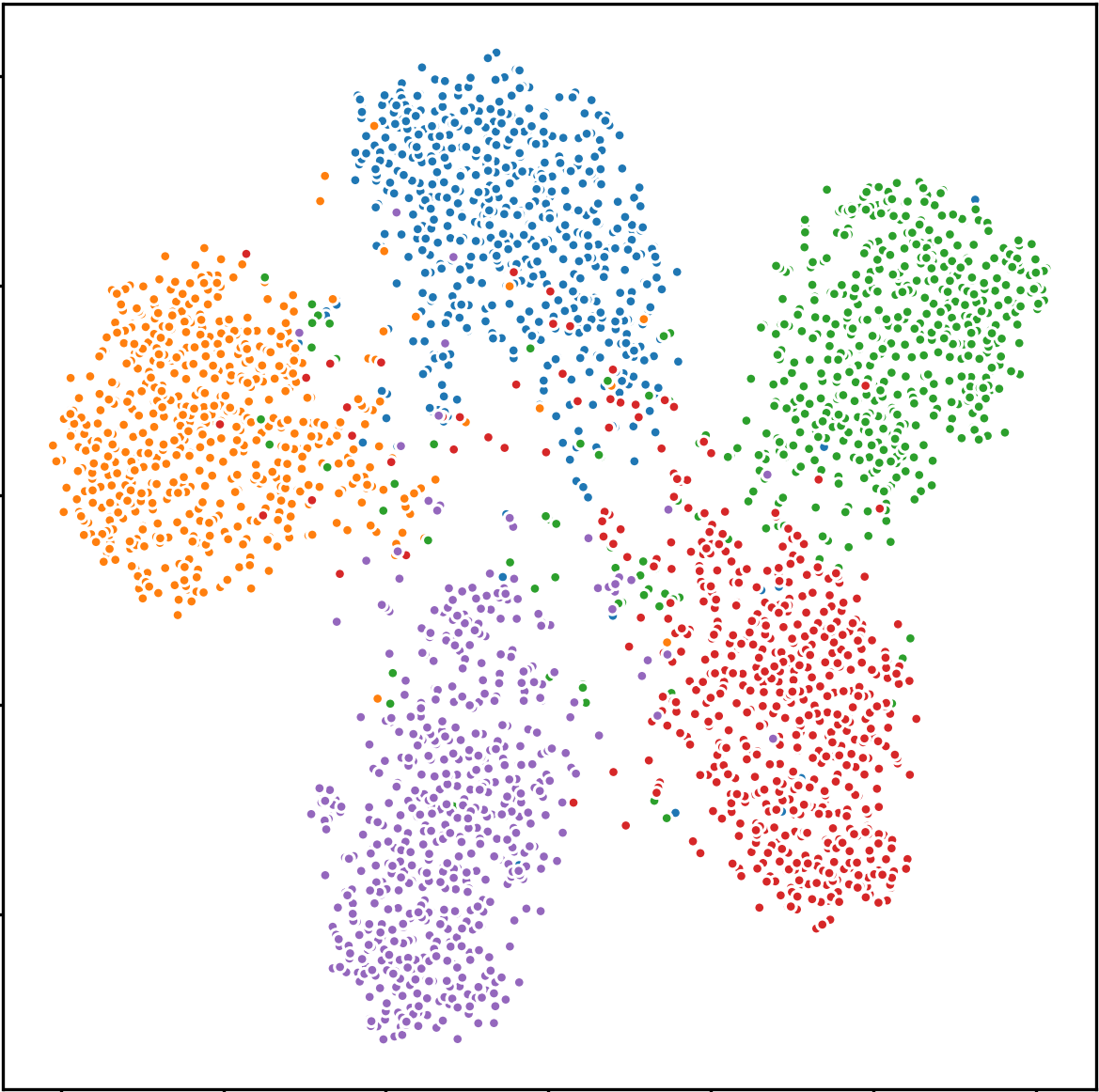}\\
\includegraphics[width=0.248\linewidth]{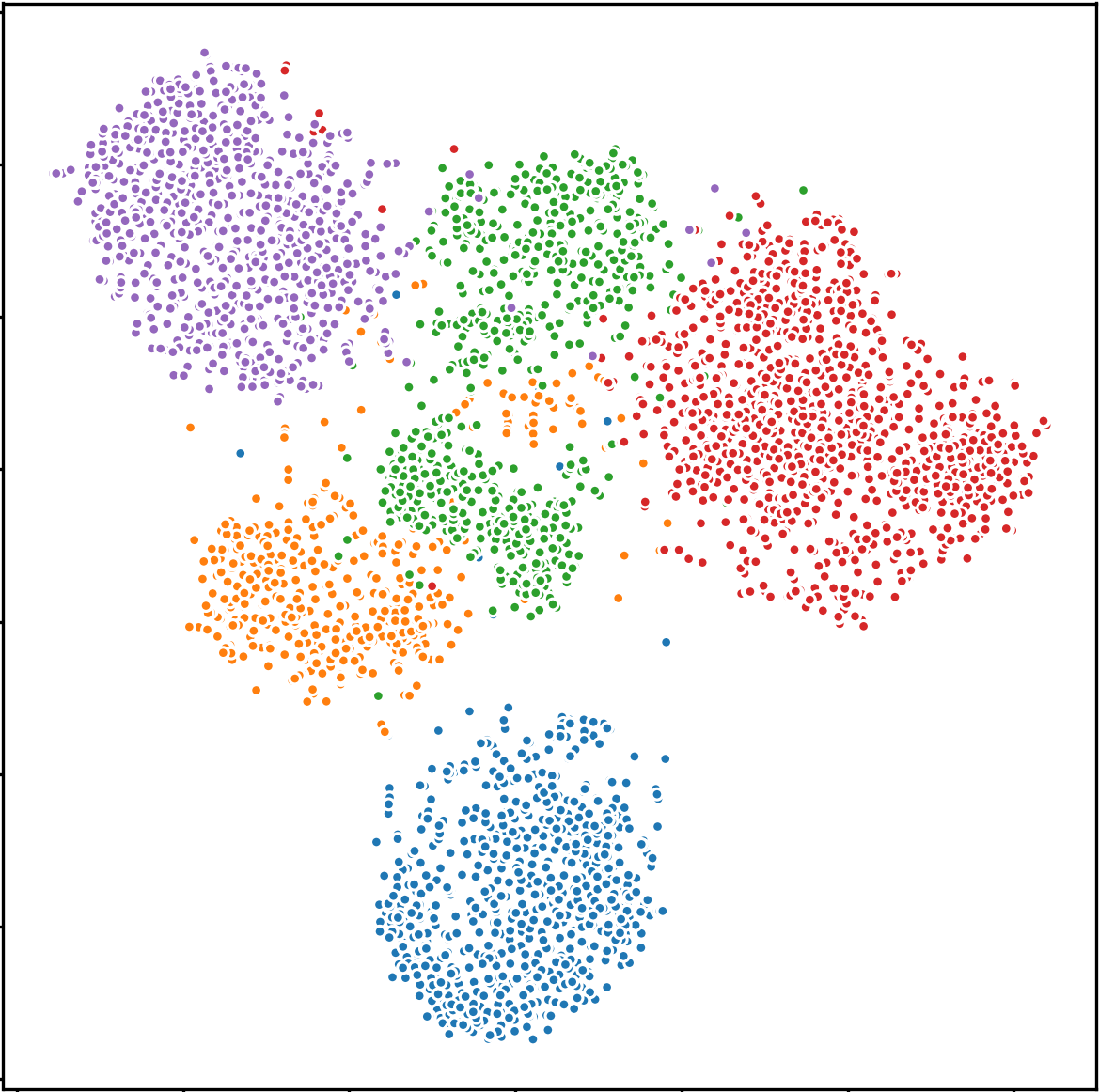}&
\includegraphics[width=0.248\linewidth]{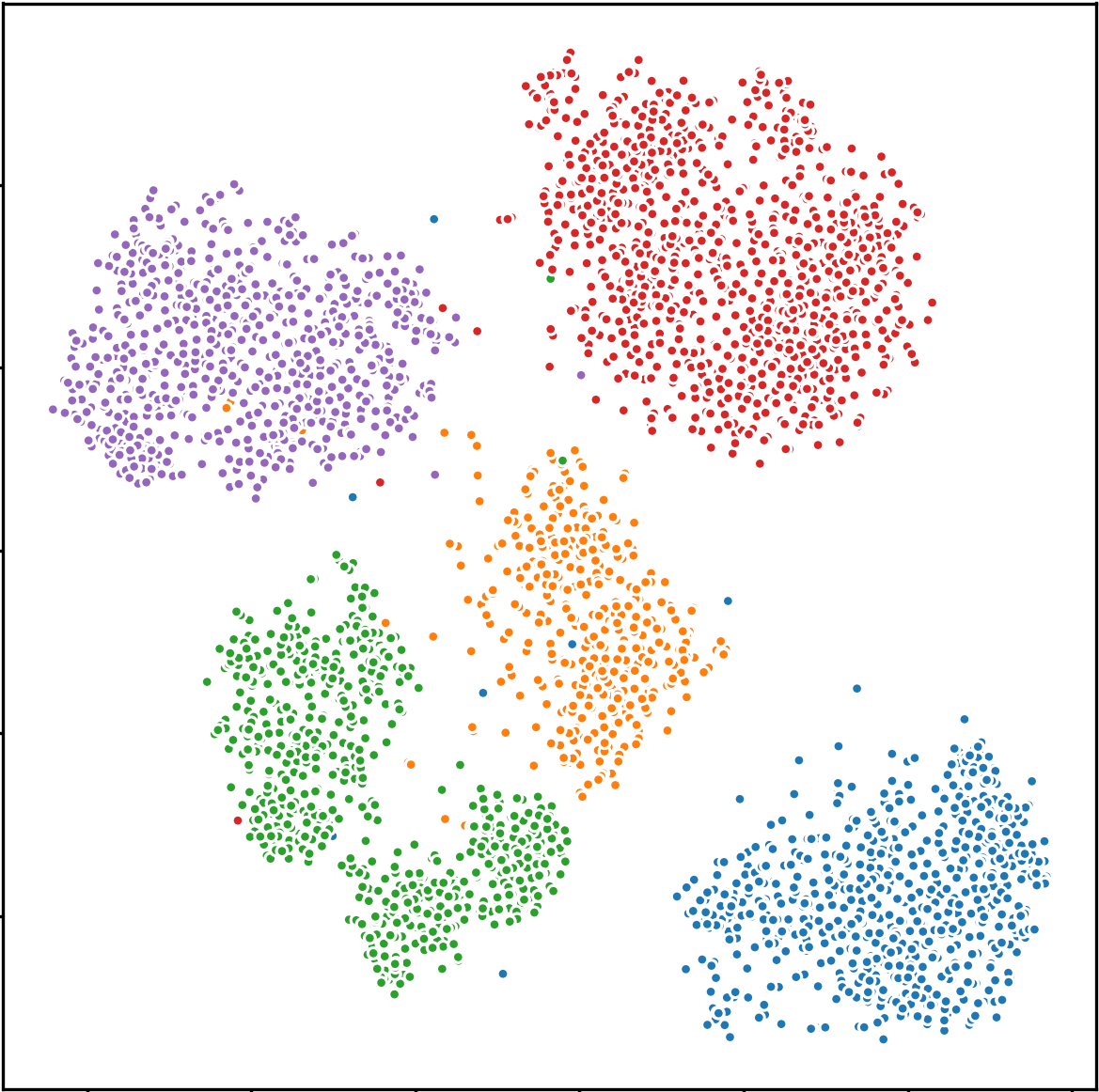}&
\includegraphics[width=0.248\linewidth]{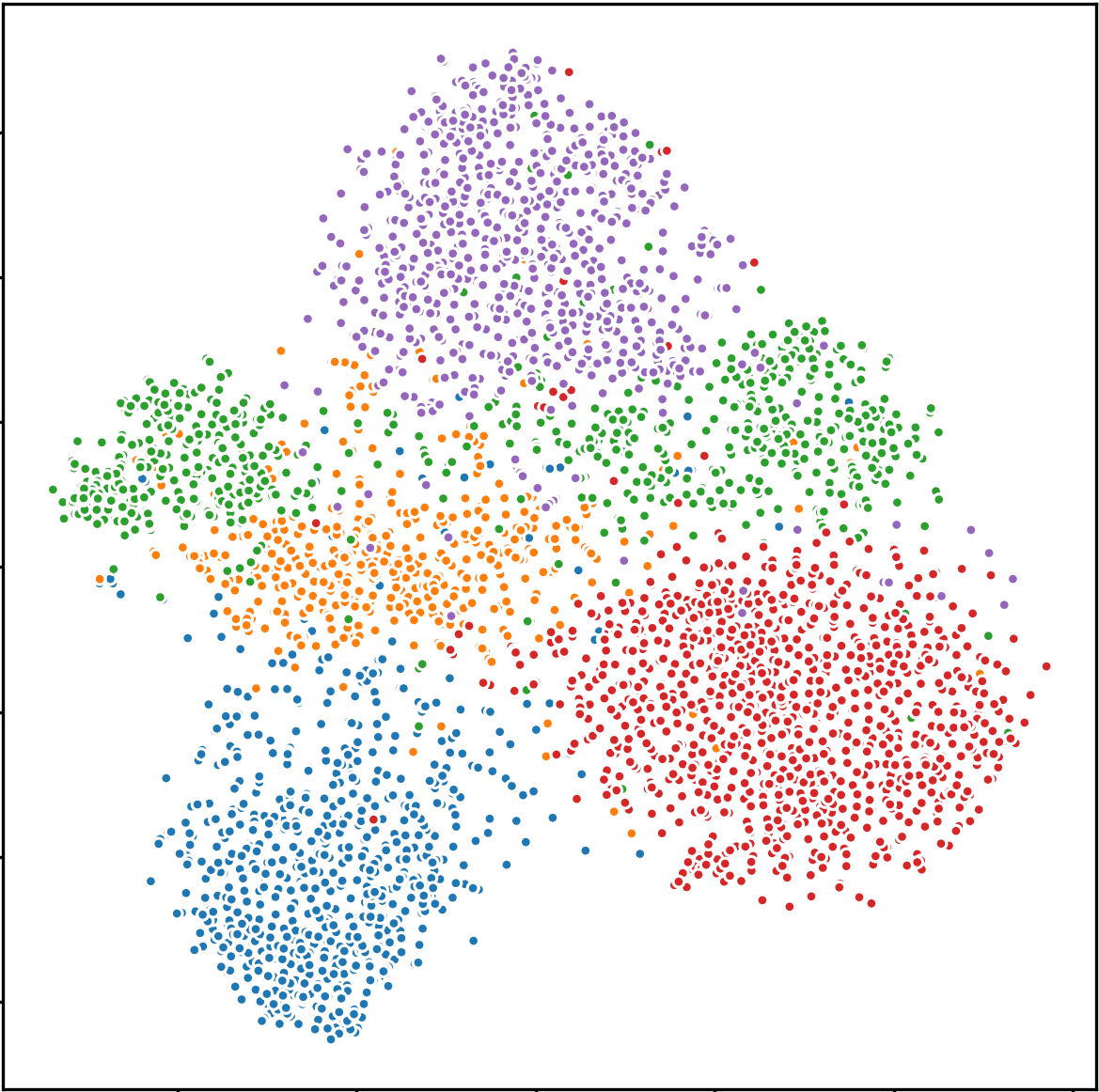}&
\includegraphics[width=0.248\linewidth]{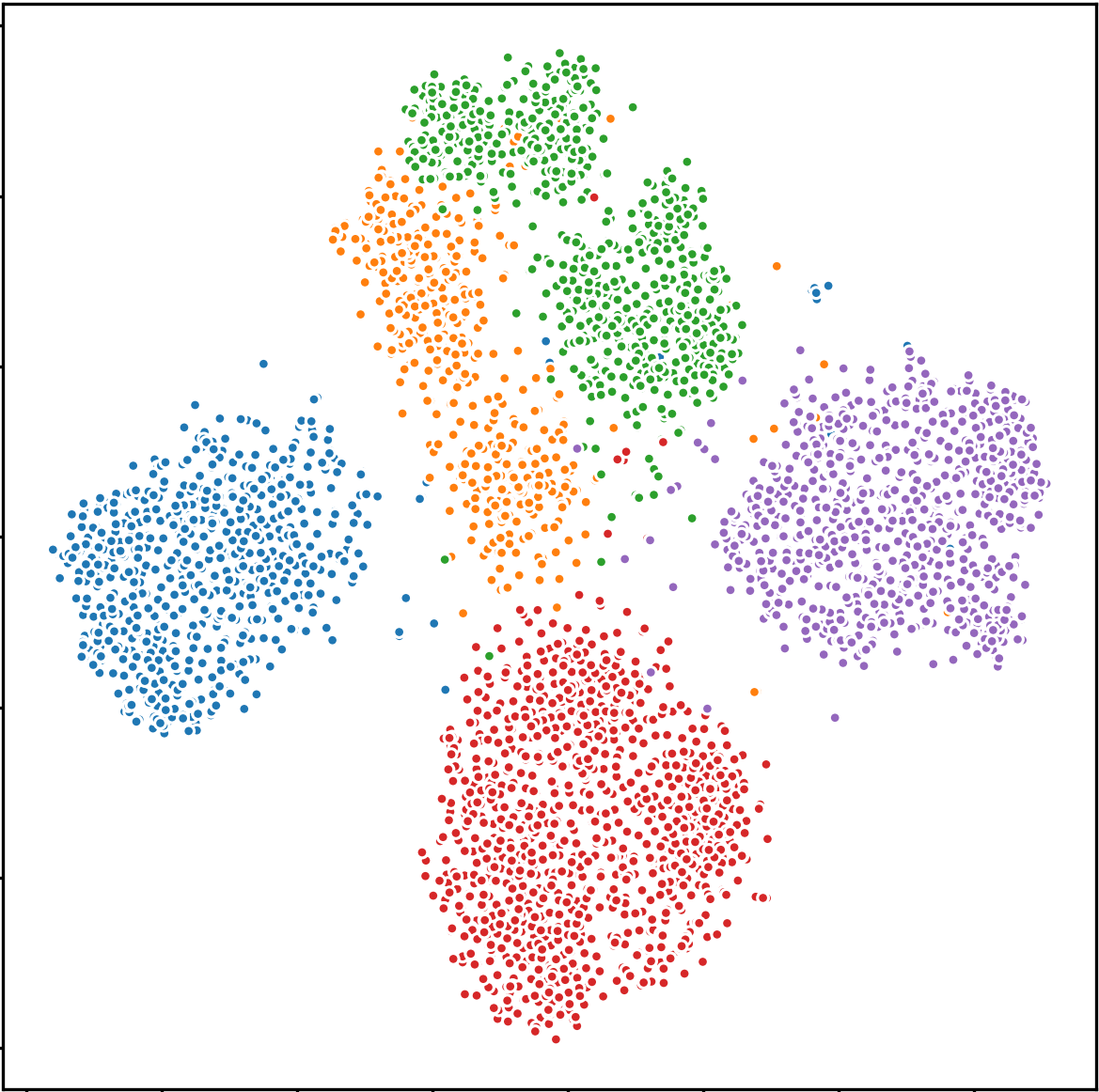}\\
\scriptsize{ResNet10+CE}&\scriptsize{ResNet10+CE$^\dagger$}&\scriptsize{ResNet10+BSR}&\scriptsize{ResNet10+BSR$^\dagger$}
\end{tabular}
\caption{\textbf{Feature embedding visualization.} The t-SNE visualizations of the feature embeddings based on the pre-trained model on the source domain (1st row) and on the target CropDisease domain (2nd row). The method with $\dagger$ is our proposed feature extractor.}
\label{fig:tsne}
\vspace{-0.2cm}
\end{figure}

\begin{table*}[!htpb]
\setlength{\tabcolsep}{1pt}
\scriptsize
\centering
\caption{\textbf{Comparison with SOTA methods}. The 5-way K-shot classification accuracy on 8 datasets with ResNet10 as the backbone. Our proposed method with CE+CE and BSR+CE losses are respectively denoted as ``NASE$^\dagger$" and ``NASE$^\ddagger$". The $(+)$ denotes that the
data augmentation and label propagation techniques are used.}
\label{tab:mainresults}
\begin{tabular}{L{42pt}C{36pt}C{36pt}C{36pt}|C{36pt}C{36pt}C{36pt}|C{36pt}C{36pt}C{36pt}|C{36pt}C{36pt}C{36pt}}
\toprule
\multirow{2}{*}{Methods} & \multicolumn{3}{c|}{ISIC}&\multicolumn{3}{c|}{EuroSAT}& \multicolumn{3}{c|}{CropDisease}&\multicolumn{3}{c}{ChestX}\\
& 5-shot & 20-shot & 50-shot &5-shot & 20-shot & 50-shot& 5-shot & 20-shot & 50-shot &5-shot & 20-shot & 50-shot \\
\midrule
Fine-tune\cite{guo2020broader}&48.11$\pm$0.64&59.31$\pm$0.48&66.48$\pm$0.56&79.08$\pm$0.61&87.64$\pm$0.47&90.89$\pm$0.36&89.25$\pm$0.51&95.51$\pm$0.31&97.68$\pm$0.21&25.97$\pm$0.41&31.32$\pm$0.45&35.49$\pm$0.45\\
NSAE$^\dagger$&54.05$\pm$0.63&66.17$\pm$0.59&71.32$\pm$0.61&83.96$\pm$0.57&92.38$\pm$0.33&95.42$\pm$0.34&93.14$\pm$0.47&98.30$\pm$0.19&99.25$\pm$0.14&27.10$\pm$0.44&35.20$\pm$0.48&38.95$\pm$0.70\\
BSR\cite{liu2020feature}&54.42$\pm$0.66&66.61$\pm$0.61&71.10$\pm$0.60&80.89$\pm$0.61&90.44$\pm$0.40& 93.88$\pm$0.31&92.17$\pm$0.45&97.90$\pm$0.22&99.05$\pm$0.14&26.84$\pm$0.44&35.63$\pm$0.54& 40.18$\pm$0.56\\
NSAE$^\ddagger$&55.27$\pm$0.62&67.28$\pm$0.61&72.90$\pm$0.55&84.33$\pm$0.55&92.34$\pm$0.35&95.00$\pm$0.26&93.31$\pm$0.42&98.33$\pm$0.18&99.29$\pm$0.14&27.30$\pm$0.42&35.70$\pm$0.47&38.52$\pm$0.71\\
\midrule
LMMPQS\cite{yeh2020large}&51.88$\pm$0.60&64.88$\pm$0.58&69.46$\pm$0.58&86.30$\pm$0.53&92.59$\pm$0.31&94.16$\pm$0.28&93.52$\pm$0.39&97.60$\pm$0.23&98.24$\pm$0.17&26.10$\pm$0.44&32.58$\pm$0.47&38.22$\pm$0.52\\
NSAE$^\dagger$(+)&54.86$\pm$0.67&66.53$\pm$0.60&72.00$\pm$0.60&87.04$\pm$0.51&93.89$\pm$0.30&\textbf{96.55$\pm$0.29}&95.65$\pm$0.35&99.10$\pm$0.16&99.67$\pm$0.12&27.58$\pm$0.47&\textbf{37.12$\pm$0.52}&40.74$\pm$0.73\\
BSR(+)&56.82$\pm$0.68&67.31$\pm$0.57&72.33$\pm$0.58&85.97$\pm$0.52&93.73$\pm$0.29&96.07$\pm$0.30&95.97$\pm$0.33&99.10$\pm$0.12&99.66$\pm$0.07&28.50$\pm$0.48&36.95$\pm$0.52&\textbf{42.32$\pm$0.53}\\
NSAE$^\ddagger$(+)&\textbf{56.85$\pm$0.67}&\textbf{67.45$\pm$0.60}&\textbf{73.00$\pm$0.56}&\textbf{87.53$\pm$0.50}&\textbf{94.21$\pm$0.29}&96.50$\pm$0.29&\textbf{96.09$\pm$0.35}&\textbf{99.20$\pm$0.14}&\textbf{99.70$\pm$0.09}&\textbf{28.73$\pm$0.45}&36.14$\pm$0.50&41.80$\pm$0.72\\
\bottomrule
\end{tabular}

\begin{tabular}{L{42pt}C{36pt}C{36pt}C{36pt}|C{36pt}C{36pt}C{36pt}|C{36pt}C{36pt}C{36pt}|C{36pt}C{36pt}C{36pt}}
\toprule
\multirow{2}{*}{Methods} & \multicolumn{3}{c|}{Car}&\multicolumn{3}{c|}{CUB}& \multicolumn{3}{c|}{Plantae}&\multicolumn{3}{c}{Places}\\
& 5-shot & 20-shot & 50-shot &5-shot &20-shot & 50-shot& 5-shot & 20-shot & 50-shot &5-shot &20-shot &50-shot \\
\midrule
Fine-tune&52.08$\pm$0.74&79.27$\pm$0.63&--&64.14$\pm$0.77&84.43$\pm$0.65&89.61$\pm$0.55&59.27$\pm$0.70&75.35$\pm$0.68&81.76$\pm$0.56&70.06$\pm$0.74&80.96$\pm$0.65&84.79$\pm$0.58\\
NSAE$^\dagger$&54.91$\pm$0.74&79.68$\pm$0.54&--&68.51$\pm$0.76&85.22$\pm$0.56&89.42$\pm$0.62&59.55$\pm$0.74&75.70$\pm$0.64&82.42$\pm$0.55&71.02$\pm$0.72&82.70$\pm$0.58&85.90$\pm$0.59\\
BSR&57.49$\pm$0.72&81.56$\pm$0.78&--&69.38$\pm$0.76&85.84$\pm$0.79&90.91$\pm$0.56&61.07$\pm$0.76&77.20$\pm$0.90&82.16$\pm$0.59&71.09$\pm$0.68&81.76$\pm$0.81&85.67$\pm$0.57\\
NSAE$^\ddagger$&58.30$\pm$0.75&82.32$\pm$0.50&--&71.92$\pm$0.77&88.09$\pm$0.48&91.00$\pm$0.79&62.15$\pm$0.77&77.40$\pm$0.65&83.63$\pm$0.60&73.17$\pm$0.72&82.50$\pm$0.59&85.92$\pm$0.56\\
\midrule
GNN-FT\cite{tseng2020cross}&44.90$\pm$0.64&--&--&66.98$\pm$0.68&--&--&53.85$\pm$0.62&--&--&\textbf{73.94$\pm$0.67}&--&--\\
NSAE$^\dagger$(+)&55.51$\pm$0.73&83.17$\pm$0.56&--&69.96$\pm$0.80&89.01$\pm$0.54&93.11$\pm$0.64&61.71$\pm$0.79
&78.58$\pm$0.64&84.64$\pm$0.76&71.86$\pm$0.72&\textbf{83.24$\pm$0.58}&86.22$\pm$0.70\\
BSR(+)&59.82$\pm$0.76&82.39$\pm$0.51&--&73.83$\pm$0.74&90.88$\pm$0.42&92.91$\pm$0.60&64.20$\pm$0.77&79.66$\pm$0.65&83.44$\pm$0.79&71.61$\pm$0.71&82.12$\pm$0.80&85.82$\pm$0.75\\
NSAE$^\ddagger$(+)&\textbf{61.11$\pm$0.79}&\textbf{85.04$\pm$0.52}&--&\textbf{76.00$\pm$0.71}&\textbf{91.08$\pm$0.42}&\textbf{95.41$\pm$0.50}&\textbf{65.66$\pm$0.78}&\textbf{81.54$\pm$0.60}&\textbf{85.99$\pm$0.72}&73.40$\pm$0.71&83.00$\pm$0.59&\textbf{86.53$\pm$0.77}\\
\bottomrule
\end{tabular}
\end{table*}

\subsection{Generalization capability analysis}

\noindent
\textbf{T-SNE visualization}
To qualitatively show the generalization capability of the feature encoder in our proposed method, we use t-SNE to visualize the feature embeddings of images from the source and target domain respectively in first row and second row in Fig.~\ref{fig:tsne}.
In each plot, we randomly select 5 classes on each domain and visualize the features of all images in these classes based on different pre-trained encoders without fine-tuning.
We use ResNet10 as encoder structure with CE (1st column) or BSR (3rd column) classification loss during the pre-training.
Our proposed methods correspond to the figures in the even columns. 
As shown in the first row in Fig.~\ref{fig:tsne}, since there are enough training examples on the source domain, all models exhibit discriminative structures. The feature embeddings based on BSR loss are more centered than the CE loss, as the eigenvalues of the feature maps are regularized during the training. Moreover, the feature embeddings based on the NSAE losses have larger within-class variations and smaller class margins, as 
the model takes classification and reconstruction tasks at the same time.
On the target domain, as shown in the second row in Fig.~\ref{fig:tsne}, we observe the opposite. In 1st and 3rd columns, features of different classes become confused with traditional pre-training. When the NSAE loss is used, the classes on the target domain becomes more separable. The within-class variations are smaller and the inter-class distance becomes larger. This suggests the better generalization capability of our proposed method.

\begin{figure}[t]
\centering
\vspace{-0.5cm}
\subfloat[ICC ratio]
{\includegraphics[width=0.5\linewidth]{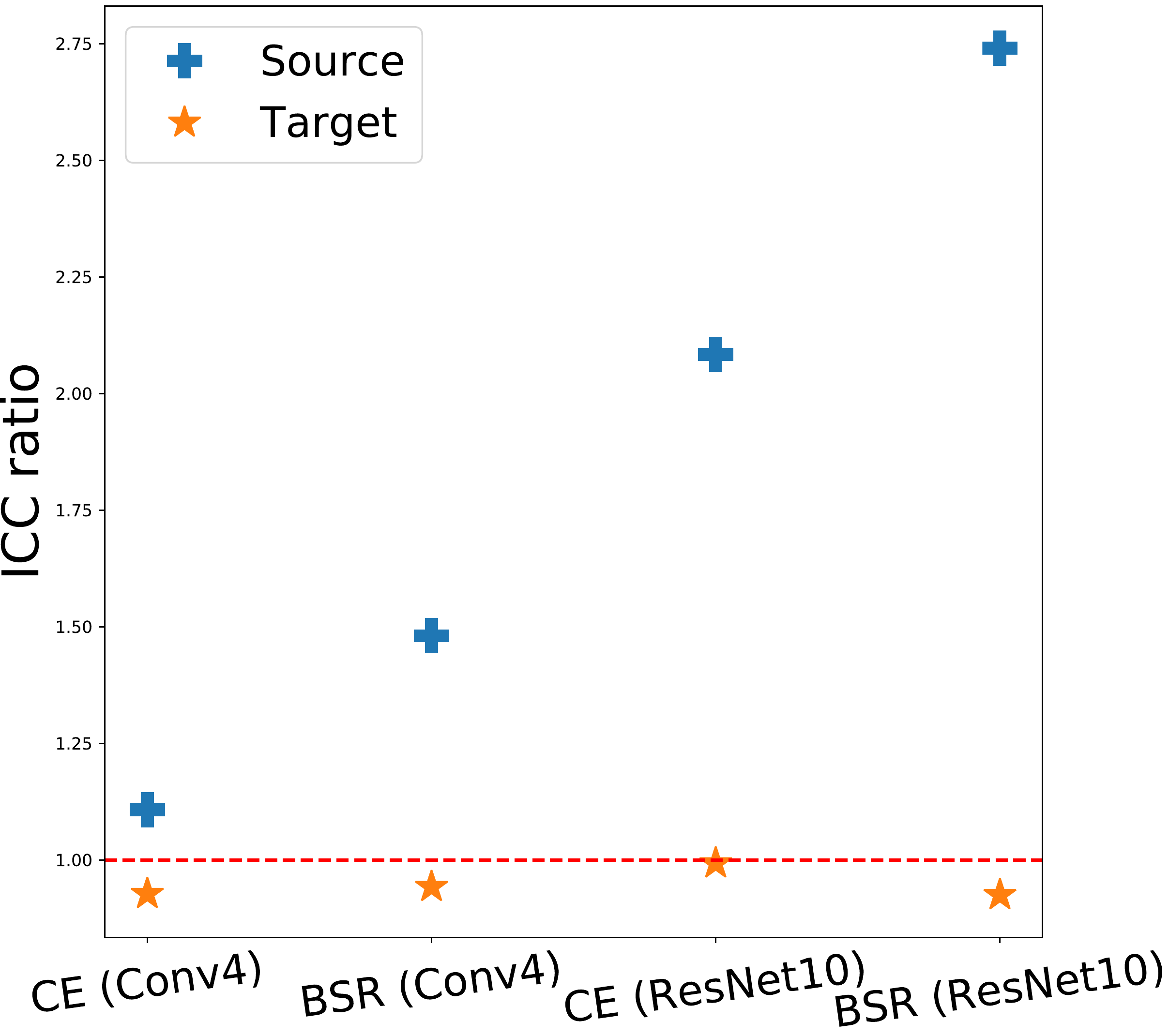}}
\subfloat[Inter-class variation ratio]
{\includegraphics[width=0.5\linewidth]{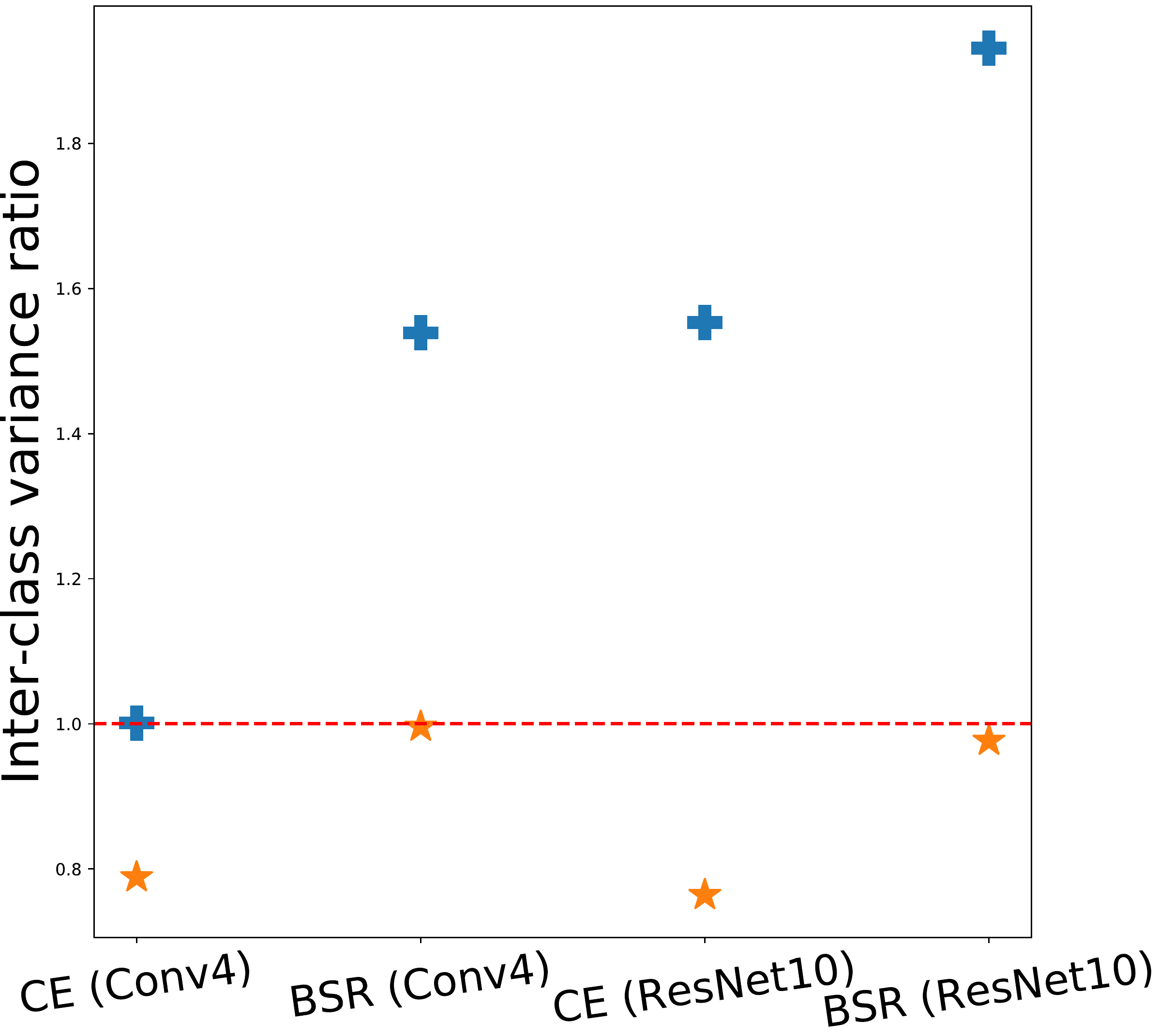}}
\caption{\textbf{ICC visualization}. The comparison of the ICC and the inter-class variance on the source domain and target datasets for different feature extractors.}
\vspace{-0.4cm}
\label{fig:icc}
\end{figure}

\noindent
\textbf{Statistical analysis of discriminability}
Moreover, we also quantitatively measures the discriminability of the feature embeddings by the intra-class correlation (ICC). The ICC is defined as the ratio of inter-class variation and the intra-class variation. Therefore, the larger the ICC, the features in different classes are more separated or the features within the same classes are more concentrated. The details of the definition of ICC are in Section A in supplementary file. We compare the ICC of the features extracted from the traditionally pre-trained encoder and that based on our proposed NSAE without fine-tuning.
We use two kinds of encoder, i.e. Conv4 and ResNet10, and two kinds of classification loss, i.e. CE and BSR, during pre-training.
This leads to four combinations denoted as CE (Conv4), BSR (Conv4), CE (ResNet10), and BSR (ResNet10). 
We take the ratio of the ICCs of the traditionally trained method and that based on our proposed method. 
The results are given in Fig.~\ref{fig:icc}(a). As shown in the figure, the ICC ratios are greater than 1 on the source domain (blue crosses) and smaller than 1 on the target domain (yellow stars) for all 4 scenarios. This suggests on the source domain, the feature extractor from our proposed method is not as discriminative as that trained with traditional methods. However, these feature extractors generalize better on the target domain.  
We similarly show the inter-class variations in Fig.~\ref{fig:icc}(b). Our proposed method shows a larger inter-class variation on the target domain, suggesting that the classes are more separable.

\subsection{Main results}
Based on the ablation study, we use the combinations of \textbf{CE+CE} and \textbf{BSR+CE} as classification losses. We use $^\dagger$ to denote method with \textbf{CE+CE} losses and $^\ddagger$ to denote method with \textbf{BSR+CE} losses, and ResNet10 is used as feature encoder to compare with the SOTAs.
Using traditional transfer learning, the CE+CE  reduces to the ``Fine-tune" method in~\cite{guo2020broader} and BSR+CE reduces to the "BSR" method in~\cite{liu2020feature}. 
Note that for these methods, since they only implement on ISIC, EuroSAT, CropDisease, and ChestX, we self-implement their models on Car, CUB, Plantae, and Places datasets with their public codebase and report the results. To further improve the performance of our model, we also used data augmentation and label propagation techniques, denoted with $(+)$ in Table~\ref{tab:mainresults}. In Table~\ref{tab:mainresults}, by comparing ``Fine-tune" with ``NASE$^\dagger$", ``BSR" with ``NASE$^\ddagger$",  we can observe that our proposed method can improve on baselines by a large margin across different unseen domains for all shots.
We attribute it to the great generalization ability of the pre-training mechanism and two-step domain adaption in the fine-tuning stage. Adding augmentation techniques can further improve the results and our proposed method performs favorably against other SOTAs across different unseen domains and evaluation settings.
\vspace{-0.05cm}
\section{Conclusion}
In this work, we propose a novel method for improving the generalization capability of the transfer learning based methods for cross-domain few-shot learning(CDFSL). We propose to train a noise-enhanced supervised autoencoder instead of a simple feature extractor on the source domain. 
Theoretical analysis shows that NSAE can largely improve the generalization capability of the feature extractor.
We also leverage the nature of the autoencoder and propose a two-step fine-tuning procedure that outperforms the past one-step fine-tune procedure.
Extensive experiments and analysis demonstrate the efficacy and generalization of our method.
Moreover, the formulation of NSAE makes it very easy to apply our proposed method to existing transfer learning based methods for CDFSL to further boost their performance.


\bibliographystyle{ieee_fullname}
\bibliography{main}
\clearpage

\appendix

\section{Discriminability Analysis of Deep Features}
Below we give the details of the definition of the Inter-class correlation (ICC)~\cite{liu2020negative,mika1999fisher}.
Let $f$ be an feature extractor and $\mathcal{D}=\mathcal{D}_1\cup \mathcal{D}_2\cup\cdots \cup \mathcal{D}_K$ where $\mathcal{D}_j= \{(x_i, y_i):y_i = j\}$ be a dataset with $K$ classes. Let $\tilde{f}(x_i) := \frac{f(x_i)}{\|f(x_i)\|_2}$ be the normalized feature extracted by the feature extractor $f$. Then the center of the images features in $j$th class is defined as
\begin{equation}
    \mu(f|\mathcal{D}_j) = |\mathcal{D}_j|^{-1}\sum_{x_i\in \mathcal{D}_j} \tilde{f}(x_i).
\end{equation}
Then the classical intra-class and inter-class variation on the full dataset $\mathcal{D}$ are defined respectively as
\begin{equation}
\begin{split}
    D_{\text{intra}}(f|\mathcal{D}) &= \frac{1}{K}\sum_{k=1}^K \left\{|\mathcal{D}_k|^{-1} \sum_{x_i\in \mathcal{D}_k}\| \tilde{f}(x_i)-\mu(\mathcal{D}_k)\|^2\right\},\\
    D_{\text{inter}}(f|\mathcal{D}) &= \frac{1}{K(K-1)}\sum_{k=1}^K \sum_{j\neq k} \|\mu(\mathcal{D}_j)-\mu(\mathcal{D}_k)\|^2.
\end{split}    
\end{equation}
The inter-class variation measures the average pairwise distances of class centers and the intra-class variation measures the within class variation of the image features.
Following ~\cite{mika1999fisher}, the intra-class correlation (ICC) is defined as
\begin{equation}
    \text{ICC}(f|\mathcal{D}) = D_{\text{inter}}(f|\mathcal{D})/D_{\text{intra}}(f|\mathcal{D}).
\end{equation}
Therefore, the ICC of a feature extractor $f$ on dataset $\mathcal{D}$ is larger when the inter-class is larger and the intra-class is smaller. The ICC can therefore measures the discriminability of a feature extractor since a good feature embedding has smaller within-class variation and larger margin across classes.

In our experiment to study the discriminability of the feature extractors, we randomly sample 5 classes and compute the ICC based on the images from these 5 classes. We repeat these procedure for $600$ times and use the average ICC as a measure for the discriminability of a feature extractor. The average ICC is computed using the same feature extractor on both the source and the target domains.

\section{Model Architecture}
In our proposed noise-enhanced supervised autoencoder, we use Conv4 and ResNet10 as the encoder structure and design the corresponding decoders. 
The decoder can be seen as a mirror mapping of the encoder which consist of deconvolutional blocks, with each block containing 2D transposed convolution operator and ReLU activation, which expand the dimension of the feature map. Before deconvolutional layers, we also add several fully connected layers that transform feature representations from encoder.
Detailed architecture and layer specifications of the decoders are shown in Table~\ref{tab:AEConv4-architecture}. 
\begin{table}[!htbp]
\footnotesize
\caption{The architecture and layer specifications of the decoder modules of the Conv4 and ResNet10 based NSAE. Linear represents fully connected layer followed by ReLU activation. Deconv-ReLU represents a ConvTranspose2d-BatchNormalization-ReLU layer. Conv-Sigmoid represents a Conv2d-BatchNormalization-Sigmoid layer.}
\begin{tabular}{c|c}
 \toprule
 {\bf Module} & {\bf Specifications} \\ \hline  
\multirow{8}{*}{Conv4} 
& Linear, 1600$\times$512\\
\cline{2-2}
& Linear, 512$\times$1600\\
\cline{2-2}
& Reshape to 64$\times$5$\times$5\\
\cline{2-2}
&2$\times$2 Deconv-ReLU, 64 filters, stride $2$, padding $0$\\ \cline{2-2}
 & 2$\times$2 Deconv-ReLU, 64 filters, stride $2$, padding $0$ \\ \cline{2-2}
 & 2$\times$2 Deconv-ReLU, 64 filters, stride $2$, padding $0$ \\ \cline{2-2}
 & 2$\times$2 Deconv-ReLU, 3 filters, stride $2$, padding $0$ \\ 
 \cline{2-2}
 & 3$\times$3 Conv-Sigmoid, 3 filters, stride $1$, padding $1$ \\ 
 \midrule
 \multirow{8}{*}
{ResNet10} & Linear, 512$\times$512\\\cline{2-2}
& Linear, 512$\times$6272\\\cline{2-2}
& Reshape to 32$\times$14$\times$14\\\cline{2-2}
&2$\times$ 2 Deconv-ReLU, 32 filters, stride $2$, padding $0$\\ \cline{2-2}
 & 2$\times$ 2 Deconv-ReLU, 32 filters, stride $2$, padding $0$ \\ \cline{2-2}
 & 2$\times$ 2 Deconv-ReLU, 64 filters, stride $2$, padding $0$ \\ \cline{2-2}
 & 2$\times$ 2 Deconv-ReLU, 64 filters, stride $2$, padding $0$ \\ 
 \cline{2-2}
 & 3$\times$3 Conv-Sigmoid, 3 filters, stride $1$, padding $1$ \\ 
 
 \bottomrule
\end{tabular}
\label{tab:AEConv4-architecture}
\end{table} 

\begin{figure}[ht]
    \centering
    \includegraphics[width=0.9\linewidth]{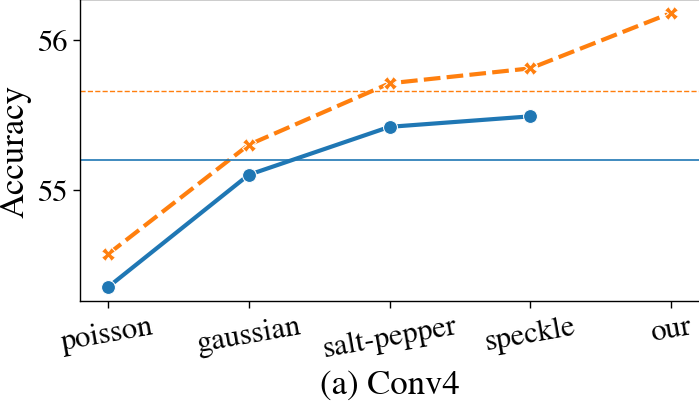}\\
    \includegraphics[width=0.9\linewidth]{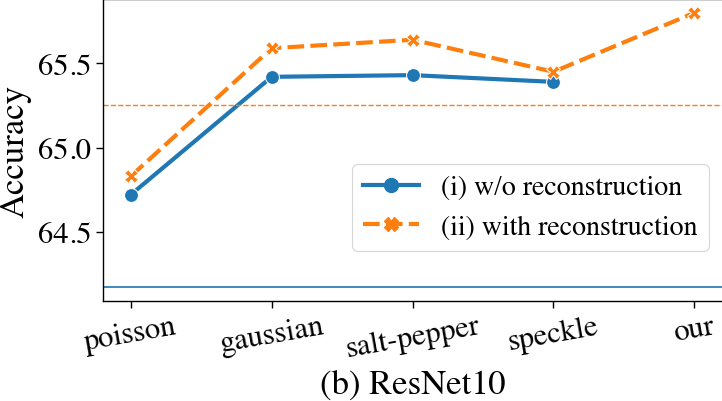}
    \caption{Ablation study with handcrafted noisy images. The two horizontal lines are baselines without noisy images.}
    \label{fig:ablation_study}
\end{figure}

\section{Ablation Study Results}
Table~\ref{tab:ablation_study} gives the detailed experiment result for the $5$-way $5$-shot ablation study on 8 datasets with various model architectures and loss functions. 
\begin{table*}[ht]
\centering
\scriptsize
\caption{\textbf{Ablation study}. The ablation study on the 5-way 5-shot support set on 8 datasets with various model architectures and loss functions. }
\begin{tabular}{l|lcccccccc}
\toprule
Encoder&Method&ISIC&EuroSAT&CropDisease&ChestX&Car&CUB&Plantae&Places\\
\midrule
\multirow{20}{*}{Conv4}&CE+CE&46.04$\pm$0.62&68.88$\pm$0.68&83.47$\pm$0.67&24.81$\pm$0.43&38.36$\pm$0.58&52.94$\pm$0.70&45.55$\pm$0.71&58.74$\pm$0.74\\ &NSAE(-)&46.64$\pm$0.62&70.19$\pm$0.64&84.89$\pm$0.60&24.86$\pm$0.41&39.88$\pm$0.64&55.08$\pm$0.70&46.98$\pm$0.69&59.54$\pm$0.69\\
&NSAE&\bf{46.65$\pm$0.63}&\bf{70.34$\pm$0.65}&\bf{85.22$\pm$0.61}&\bf{25.02$\pm$0.42}&\bf{39.90$\pm$0.62}&\bf{55.35$\pm$0.67}&\bf{47.18$\pm$0.75}&\bf{59.59$\pm$0.69}\\
&SAE&46.36$\pm$0.61&68.88$\pm$0.65&83.57$\pm$0.62&24.88$\pm$0.41&37.94$\pm$0.59&52.74$\pm$0.65&44.79$\pm$0.67&58.34$\pm$0.71\\
&SAE(*)&44.88$\pm$0.60&70.10$\pm$0.68&83.38$\pm$0.64&25.00$\pm$0.41&37.29$\pm$0.60&53.64$\pm$0.70&44.60$\pm$0.65&59.12$\pm$0.72\\
\cmidrule{2-10}
&BSR+CE&48.78$\pm$0.64&69.34$\pm$0.68&85.88$\pm$0.61&25.41$\pm$0.42&42.54$\pm$0.70&60.16$\pm$0.73&\bf{50.85$\pm$0.78}&62.38$\pm$0.77\\ 
&NSAE(-)&49.32$\pm$0.59&71.84$\pm$0.66&86.86$\pm$0.59&25.59$\pm$0.42&42.54$\pm$0.64&60.00$\pm$0.71&50.00$\pm$0.73&63.36$\pm$0.72\\
&NSAE&\bf{49.34$\pm$0.59}&72.00$\pm$0.65&\bf{86.87$\pm$0.59}&\bf{25.62$\pm$0.42}&\bf{42.56$\pm$0.65}&\bf{60.18$\pm$0.72}&49.48$\pm$0.73&\bf{63.40$\pm$0.72}\\
&SAE&48.68$\pm$0.63&\bf{72.17$\pm$0.69}&86.23$\pm$0.60&25.31$\pm$0.41&42.38$\pm$0.68&60.10$\pm$0.76&48.29$\pm$0.74&62.12$\pm$0.73\\
&SAE(*)&47.25$\pm$0.58&70.52$\pm$0.67&85.14$\pm$0.62&25.45$\pm$0.40&40.81$\pm$0.63&59.54$\pm$0.74&47.70$\pm$0.70&62.46$\pm$0.75\\
\cmidrule{2-10}
&CE+D&50.54$\pm$0.66&76.16$\pm$0.64&89.65$\pm$0.55&24.07$\pm$0.41&44.26$\pm$0.70&58.61$\pm$0.82&52.47$\pm$0.74&61.81$\pm$0.74\\
&NSAE(-)&50.94$\pm$0.63&77.70$\pm$0.70&90.06$\pm$0.52&\textbf{24.46$\pm$0.40}&43.96$\pm$0.68&\textbf{60.00$\pm$0.82}&53.26$\pm$0.80&62.26$\pm$0.73\\
&NSAE&\bf{50.95$\pm$0.63}&\bf{77.77$\pm$0.64}&\textbf{90.11$\pm$0.52}&24.29$\pm$0.40&\textbf{44.28$\pm$0.72}&59.90$\pm$0.78&\textbf{53.36$\pm$0.78}&\textbf{62.42$\pm$0.72}\\
&SAE&50.62$\pm$0.65&74.92$\pm$0.64&88.11$\pm$0.59&23.52$\pm$0.41&42.45$\pm$0.70&57.04$\pm$0.81&51.51$\pm$0.80&61.08$\pm$0.74\\
&SAE(*)&49.94$\pm$0.66&77.24$\pm$0.70&88.31$\pm$0.56&24.01$\pm$0.40&42.12$\pm$0.68&57.15$\pm$0.82&51.77$\pm$0.82&61.40$\pm$0.78\\
\cmidrule{2-10}
&BSR+D&\textbf{50.06$\pm$0.65}&75.74$\pm$0.67&87.71$\pm$0.56&\textbf{23.66$\pm$0.40}&41.11$\pm$0.77&58.81$\pm$0.81&51.35$\pm$0.81&60.51$\pm$0.80\\ 
&NSAE(-)&49.95$\pm$0.67&76.96$\pm$0.70&87.70$\pm$0.57&23.60$\pm$0.41&41.05$\pm$0.72&58.32$\pm$0.81&51.74$\pm$0.84&60.34$\pm$0.81\\
&NSAE&49.98$\pm$0.67&\textbf{77.00$\pm$0.69}&\textbf{87.71$\pm$0.58}&23.61$\pm$0.41&\textbf{41.80$\pm$0.72}&\textbf{59.42$\pm$0.82}&\textbf{51.80$\pm$0.84}&\textbf{60.92$\pm$0.85}\\
&SAE&49.77$\pm$0.68&75.58$\pm$0.69&87.67$\pm$0.57&23.35$\pm$0.41&41.75$\pm$0.75&58.34$\pm$0.81&50.92$\pm$0.81&60.25$\pm$0.81\\
&SAE(*)&49.46$\pm$0.68&76.17$\pm$0.70&86.50$\pm$0.60&23.23$\pm$0.39&40.16$\pm$0.70&58.26$\pm$0.79&50.70$\pm$0.83&60.86$\pm$0.84\\
\midrule
\multirow{20}{*}{ResNet10}&CE+CE&51.28$\pm$0.62&82.51$\pm$0.58&92.45$\pm$0.45&26.50$\pm$0.43&52.08$\pm$0.72&64.14$\pm$0.77&59.27$\pm$0.70&70.06$\pm$0.74\\
&NSAE(-)&53.52$\pm$0.62&83.83$\pm$0.56&93.14$\pm$0.47&26.69$\pm$0.44&53.49$\pm$0.72&67.60$\pm$0.73&59.70$\pm$0.74&70.74$\pm$0.71\\
&NSAE&\textbf{54.05$\pm$0.63}&\textbf{83.96$\pm$0.57}&\textbf{93.14$\pm$0.47}&\textbf{27.10$\pm$0.44}&\textbf{54.91$\pm$0.74}&\textbf{68.51$\pm$0.76}& \textbf{59.80$\pm$0.74}&\textbf{71.84$\pm$0.72}\\
&SAE&52.28$\pm$0.63&83.78$\pm$0.55&93.01$\pm$0.42&26.05$\pm$0.45&53.54$\pm$0.71&64.27$\pm$0.75&59.87$\pm$0.73&70.82$\pm$0.72\\
&SAE(*)&52.11$\pm$0.65&83.50$\pm$0.55&93.05$\pm$0.47&26.37$\pm$0.45&54.26$\pm$0.70&66.62$\pm$0.75&59.62$\pm$0.75&71.40$\pm$0.67\\
\cmidrule{2-10}
&BSR+CE&54.42$\pm$0.66&80.89$\pm$0.61&92.17$\pm$0.45&26.84$\pm$0.44&57.49$\pm$0.72&69.38$\pm$0.76&61.07$\pm$0.76&71.09$\pm$0.68\\
&NSAE(-)&55.27$\pm$0.62&84.19$\pm$0.54&92.92$\pm$0.47&27.23$\pm$0.45&\textbf{58.35$\pm$0.76}&71.30$\pm$0.75&61.92$\pm$0.76&71.76$\pm$0.74\\
&NSAE&\textbf{55.88$\pm$0.64}&\textbf{84.33$\pm$0.55}&\textbf{93.31$\pm$0.42}&\textbf{27.30$\pm$0.42}&58.30$\pm$0.75&\textbf{71.92$\pm$0.77}&\textbf{62.18$\pm$0.77}&\textbf{73.17$\pm$0.72}\\
&SAE&54.48$\pm$0.65&84.10$\pm$0.54&92.92$\pm$0.47&27.20$\pm$0.45&58.30$\pm$0.76&71.30$\pm$0.75&61.92$\pm$0.76&71.76$\pm$0.74\\
&SAE(*)&54.73$\pm$0.68&83.90$\pm$0.55&93.02$\pm$0.46&26.74$\pm$0.43&57.60$\pm$0.71&71.50$\pm$0.75&62.20$\pm$0.78&72.99$\pm$0.67\\
\cmidrule{2-10}
&CE+D&51.62$\pm$0.66&83.72$\pm$0.59&93.22$\pm$0.41&26.23$\pm$0.44&55.12$\pm$0.76&66.56$\pm$0.78&59.09$\pm$0.76&72.81$\pm$0.73\\
&NSAE(-)&54.31$\pm$0.68&83.77$\pm$0.62&93.54$\pm$0.40&26.98$\pm$0.44&55.67$\pm$0.78&67.17$\pm$0.76&59.46$\pm$0.75&72.90$\pm$0.72\\
&NSAE&\textbf{54.41$\pm$0.63}&\textbf{83.78$\pm$0.56}&\textbf{93.65$\pm$0.40}&\textbf{27.25$\pm$0.44}&\textbf{55.78$\pm$0.73}&\textbf{67.64$\pm$0.76}&\textbf{59.74$\pm$0.75}&\textbf{73.25$\pm$0.73}\\
&SAE&52.64$\pm$0.67&83.13$\pm$0.63&93.44$\pm$0.41&26.34$\pm$0.44&55.44$\pm$0.74&65.08$\pm$0.76&59.70$\pm$0.78&73.13$\pm$0.71\\
&SAE(*)&51.37$\pm$0.66&83.04$\pm$0.63&92.53$\pm$0.42&26.44$\pm$0.42&55.00$\pm$0.73&65.13$\pm$0.81&59.46$\pm$0.79&73.20$\pm$0.67\\
\cmidrule{2-10}
&BSR+D&52.85$\pm$0.65&80.13$\pm$0.65&91.20$\pm$0.48&\textbf{26.80$\pm$0.45}&54.99$\pm$0.74&68.15$\pm$0.84&58.26$\pm$0.77&71.97$\pm$0.72\\
&NSAE(-)&53.74$\pm$0.67&82.19$\pm$0.64&92.22$\pm$0.47&26.79$\pm$0.45&55.90$\pm$0.77&68.32$\pm$0.81&60.25$\pm$0.77&73.28$\pm$0.72\\
&NSAE&\textbf{54.42$\pm$0.64}&\textbf{82.79$\pm$0.62}&\textbf{92.45$\pm$0.45}&26.69$\pm$0.45&\textbf{55.92$\pm$0.72}&\textbf{68.46$\pm$0.82}&\textbf{60.40$\pm$0.78}&\textbf{73.33$\pm$0.71}\\
&SAE&51.84$\pm$0.65&80.02$\pm$0.69&91.95$\pm$0.45&26.52$\pm$0.42&55.90$\pm$0.77&66.64$\pm$0.79&59.20$\pm$0.80&72.48$\pm$0.76\\
&SAE(*)&53.08$\pm$0.67&81.77$\pm$0.64&91.63$\pm$0.46&26.58$\pm$0.45&54.87$\pm$0.78&67.97$\pm$0.83&58.61$\pm$0.79&73.20$\pm$0.67\\
\bottomrule
\end{tabular}
\label{tab:ablation_study}
\end{table*}
We use four kinds of combinations of the classification loss functions for pre-training and fine-tuning, i.e. CE+CE, BSR+CE, CE+D, and BSR+D. Meanwhile, we respectively test with Conv4 and ResNet10 as backbone of feature encoder. In the table, CE+CE, BSR+CE, CE+D, and BSR+D denote using single feature extractor with different loss functions combinations. SAE denotes that we use auto-encoder but do not further feed in the reconstructed images for classification during the pre-training. SAE(*) denotes that we double the weight on the classification loss of original images as if the auto-encoder works perfectly that the reconstructed images are identical to original images. NSAE(-) denotes using our proposed pre-training strategy but using one-step fine-tuning.

\section{Comparison with Handcrafted Noise}
The reconstructed images during the pre-training stage can be viewed as noisy inputs to improve the model generalization capability. 
Can the model generalization capability be improved if we use images with handcrafted noise instead of reconstructed images?
To answer this question, we compare the performance of our proposed method with that when images with handcrafted noise are used as data augmentation during pre-training. 
In our experiment, we consider the following four kinds of handcrafted noise: Gaussian, salt-pepper, Poisson, and speckle. 
We use the \emph{skimage} package~\cite{van2014scikit} in python to add handcrafted noise to source images. The parameter values for the noise generation are given in Table~\ref{tab:skimage_noise}.
\begin{table}[ht]
\caption{\textbf{Handcrafted Noise Configuration}. The parameters for adding noise to the images.}
\centering
\begin{tabular}{ll}
\toprule
Noise type & Parameter values\\
\midrule
Gaussian   & mode=`gaussian', mean=0, var=0.1\\
salt-pepper & mode=`s\&p', salt\_vs\_pepper=0.5\\
Poisson & mode=`poisson'\\
speckle & mode=`speckle', mean=0, var=0.05\\
\bottomrule
\end{tabular}
\label{tab:skimage_noise}
\end{table}
We use BSR+CE loss combination and consider the following two settings during pre-training: (a) only use the encoder and feed in both source and handcrafted noisy images for classification; (b) add a decoder to (a) with reconstruction loss, though the reconstructed images are not used for classification.
The rest of the hyper-parameter values are the same as that given in Section 4.1 in the main paper.
The results averaged over 8 datasets are shown in Fig.~\ref{fig:ablation_study}.

It can be seen from the Fig.~\ref{fig:ablation_study} that 
\begin{enumerate}
    \item regardless of the noise type, using auto-encoder scheme with reconstruction loss helps improve the generalization capability owing to regularization effect from decoder, which shows the advantage of our model on top of simple data augmentation;
    \item adding handcrafted noise may not improve the accuracy, but our design consistently improves the accuracy and surpasses all results with handcrafted noise. 
\end{enumerate}

\end{document}